\documentclass[letterpaper]{article} % DO NOT CHANGE THIS
\usepackage{aaai2026}  % DO NOT CHANGE THIS
\usepackage{times}  % DO NOT CHANGE THIS
\usepackage{helvet}  % DO NOT CHANGE THIS
\usepackage{courier}  % DO NOT CHANGE THIS
\usepackage[hyphens]{url}  % DO NOT CHANGE THIS
\usepackage{graphicx} % DO NOT CHANGE THIS

\usepackage{natbib}  % DO NOT CHANGE THIS AND DO NOT ADD ANY OPTIONS TO IT
\usepackage{caption} % DO NOT CHANGE THIS AND DO NOT ADD ANY OPTIONS TO IT
\usepackage{algorithm}
\usepackage{algorithmic}
\usepackage{amsmath}
\usepackage{booktabs}
\usepackage{multirow}
\usepackage{graphicx}
\usepackage{amsfonts}
\usepackage{caption}                 % 放在导言区
\usepackage{placeins}
\usepackage{newfloat}
\usepackage{listings}
\DeclareCaptionStyle{ruled}{labelfont=normalfont,labelsep=colon,strut=off} % DO NOT CHANGE THIS
\floatstyle{ruled}
\newfloat{listing}{tb}{lst}{}
\floatname{listing}{Listing}
\title{SAVER: Mitigating Hallucinations in Large Vision-Language Models via Style-Aware Visual Early Revision}
\author{
    Zhaoxu Li\textsuperscript{\rm 1,\rm 2},
    Chenqi Kong\textsuperscript{\rm 2}\thanks{Corresponding author},
    Yi Yu\textsuperscript{\rm 2},
    Qiangqiang Wu\textsuperscript{\rm 3},
    Xinghao Jiang\textsuperscript{\rm 4},
    Ngai-Man Cheung\textsuperscript{\rm 5},
    Bihan Wen\textsuperscript{\rm 2},
    Alex Kot\textsuperscript{\rm 2,\rm 6},
    Xudong Jiang\textsuperscript{\rm 2}
}

\affiliations{
    \textsuperscript{\rm 1}ROSE Lab, Interdisciplinary Graduate Programme, Nanyang Technological University, Singapore \\
    \textsuperscript{\rm 2}ROSE Lab, School of Electrical and Electronic Engineering, Nanyang Technological University, Singapore\\
    \textsuperscript{\rm 3}City University of Hong Kong, Hong Kong SAR\\
    \textsuperscript{\rm 4}Shanghai Jiao Tong University, China\\  
    \textsuperscript{\rm 5}Singapore University of Technology and Design, Singapore\\
    \textsuperscript{\rm 6}VinUniversity, Hanoi, Vietnam \\
    \{zhaoxu001, yuyi0010\}@e.ntu.edu.sg, qiangqwu2-c@my.cityu.edu.hk, xhjiang@sjtu.edu.cn, ngaiman\_cheung@sutd.edu.sg, \{chenqi.kong, bihan.wen, eackot, exdjiang\}@ntu.edu.sg
}

\usepackage{bibentry}
\begin{document}

\maketitle

\begin{abstract}
Large Vision-Language Models (LVLMs) recently achieve significant breakthroughs in understanding complex visual-textual contexts. However, hallucination issues still limit their real-world applicability. Although previous mitigation methods effectively reduce hallucinations in photographic images, they largely overlook the potential risks posed by stylized images, which play crucial roles in critical scenarios such as game scene understanding, art education, and medical analysis. In this work, we first construct a dataset comprising photographic images and their corresponding stylized versions with carefully annotated caption labels. We then conduct head-to-head comparisons on both discriminative and generative tasks by benchmarking 13 advanced LVLMs on the collected datasets. Our findings reveal that stylized images tend to induce significantly more hallucinations than their photographic counterparts. To address this issue, we propose \textbf{S}tyle-\textbf{A}ware \textbf{V}isual \textbf{E}arly \textbf{R}evision (\textbf{SAVER}), a novel mechanism that dynamically adjusts LVLMs' final outputs based on the token-level visual attention patterns, leveraging early-layer feedback to mitigate hallucinations caused by stylized images. Extensive experiments demonstrate that SAVER achieves state-of-the-art performance in hallucination mitigation across various models, datasets, and tasks.
\end{abstract}

\begin{links}
    \link{Extended version}{https://www.arxiv.org/abs/2508.03177}
\end{links}
\section{Introduction}

\label{sec:intro}
Large Vision-Language Models (LVLMs)~\cite{openai2023gpt4, chen2023shikra, yin2024survey, zhao2024harmonizing} have achieved remarkable successes in a plethora of applications in the past few years \cite{li2022blip, liu2023visual, zhu2023minigpt}. However, the phenomenon of hallucination can cause severe consequences in some practical scenarios, such as medical analyzes \cite{hu2023advancing, wang2023chatcad}, autonomous driving \cite{chen2024driving, liu2023llm}, and human-computer interaction \cite{brie2023evaluating}. These issues raise pressing security concerns and may cause unintended harm to the public, significantly hindering the real-world deployment of LVLMs.

To mitigate hallucinations, existing approaches can be broadly categorized into two groups: instruction tuning methods \cite{sun2023aligning, xing2024mitigating} and decoding-based methods \cite{chuang2023dola, jiang2024devils}. The former requires retraining LVLMs using curated datasets and tuning strategies, which effectively reduce hallucinations while introducing expensive computational costs. The latter employs post-hoc correction mechanisms, adjusting logit scores at each generative step to encourage LVLMs to focus more on the corresponding visual content. While these methods have shown promising results for photographic images, they suffer from significant performance drops when applied to stylized images (e.g., game and sketch).

\begin{figure}
\begin{center}
\includegraphics[width=0.95\linewidth]{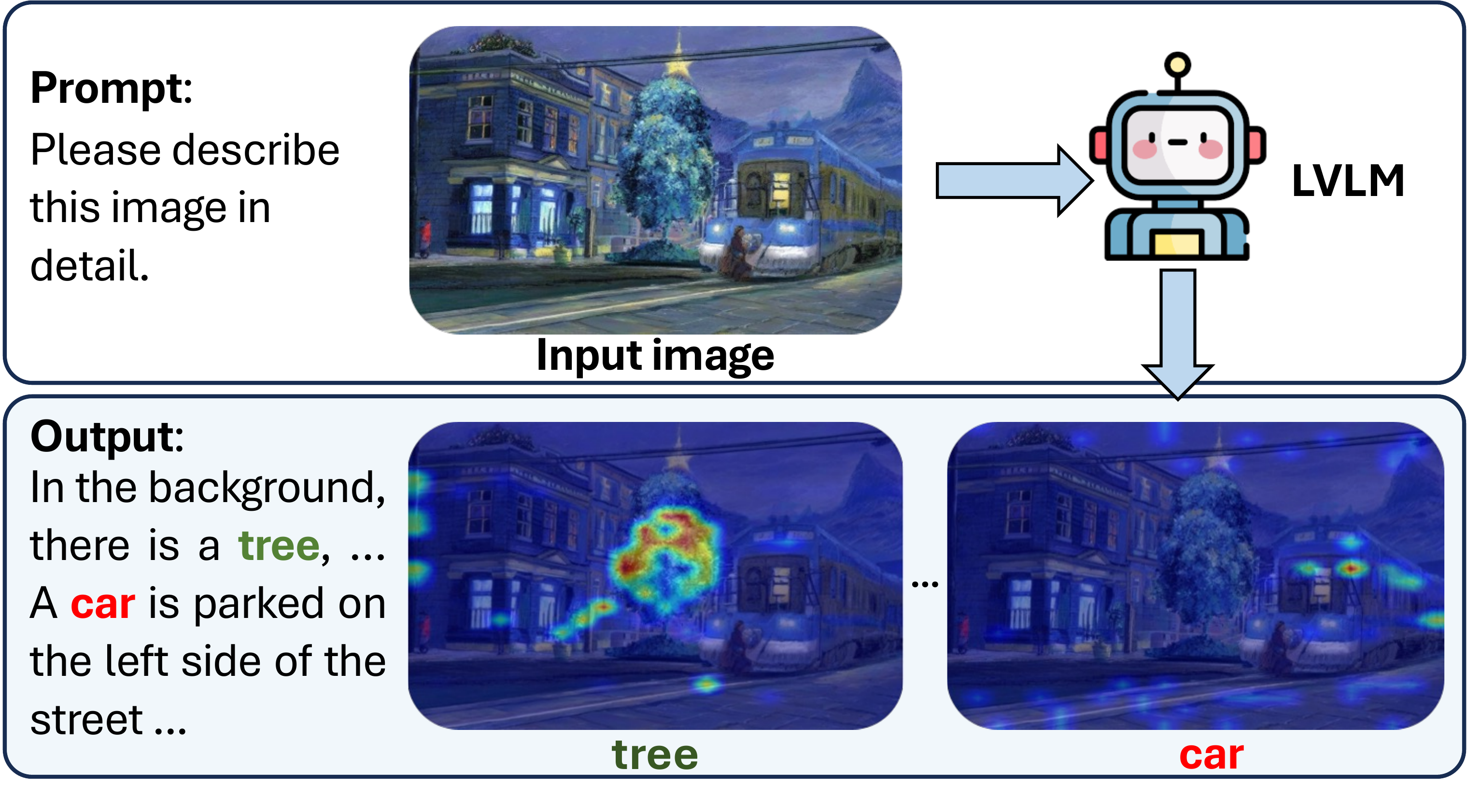}
\end{center}
   \caption{Correlation map between image tokens and generated tokens. {Red}: hallucinated tokens, showing sparse, low-confidence correlations. {Green}: real tokens, showing concentrated clusters over the corresponding regions. }
\label{fig:teaser}
\end{figure}

Stylized images play a critical role in various applications, including art education, medical analysis, and criminal forensics. For instance, LVLMs can interpret forensic sketches to retrieve matches from photographic databases or generate suspect profiles. Minimizing hallucinations in such scenarios is essential to enhance the reliability of these models. However, a comprehensive benchmark for assessing LVLM performance on stylized images remains unavailable. To fill this gap, we construct a dataset comprising paired captions and both photographic and style-transferred images generated using state-of-the-art methods across five stylistic domains. We evaluate 13 advanced LVLMs and observe that stylized images significantly increase hallucination rates compared to their original photographic counterparts.
 
To investigate the underlying causes, we conduct an empirical analysis by measuring the correlations between stylized image representations and the generated tokens. Fig.~\ref{fig:teaser} visualizes the representational patterns of real and hallucinated objects in stylized images. We observe that real object representations exhibit concentrated, high-confidence activation regions, while the hallucinated object displays sparse, low-confidence distributions. This insight naturally inspires the design of a decoding strategy that encourages the model to focus more precisely on the relevant visual regions.

We propose \textbf{SAVER}: \textbf{S}tyle-\textbf{A}ware \textbf{V}isual \textbf{E}arly \textbf{R}evision, a training-free hallucination mitigation strategy designed to address hallucination issues in stylized images. Due to the strong knowledge priors of language models, LVLMs tend to progressively suppress visual information in later model layers \cite{wang2024mllm, jiang2024devils}. To counteract this, SAVER dynamically searches the optimal preceding layers with highly concentrated representational patterns to refine the output token logits. As a plug-and-play decoding strategy, SAVER can be seamlessly integrated into various LVLMs and significantly reduces hallucination rates for stylized images without requiring additional training.

Our key contributions are: (1) To the best of our knowledge, we are the first to construct a captioning dataset specifically for stylized images. We further establish a benchmark using 13 advanced LVLMs and find that stylized images tend to produce more hallucinations; (2) Compared to hallucinated tokens, we demonstrate that the correlation patterns between correct tokens and input visual contents exhibit denser visual activations in early layers. Based on this insight, we propose SAVER, a training-free decoding strategy that dynamically corrects generated tokens by leveraging high-confidence activations from earlier layers; (3) Extensive experimental results show that SAVER consistently outperforms previous methods, effectively mitigating hallucinations across a variety of models, datasets, and tasks.

\section{Related Work}
\label{sec:RelatedWork}
\textbf{Large Vision-Language Models (LVLMs).} Large Language Models (LLMs) such as LLaMA~\cite{touvron2023llama} and Vicuna~\cite{Vicuna} have achieved remarkable advancements. The rapid development of LVLMs has significantly enhanced the ability of foundation models to interpret and reason about visual content. Early LVLMs, including BLIP~\cite{li2022blip} and LLaVA~\cite{liu2023visual}, extended LLMs to handle image understanding and reasoning tasks. To bridge the modality gap between vision and language, various approaches have been adopted, such as linear projection layers (e.g., LLaVA~\cite{liu2023visual}, MiniGPT-4~\cite{zhu2023minigpt}), Q-former modules (e.g., BLIP-2~\cite{li2023blip2}, InstructBLIP~\cite{instructblip}), and cross-attention mechanisms (e.g., Flamingo~\cite{alayrac2022flamingo}, OpenFlamingo~\cite{awadalla2023openflamingo}). More recent models, such as GPT-4o~\cite{openai2023gpt4} and Gemini~\cite{gemini_llm}, have demonstrated exceptional capabilities in visual reasoning and understanding. Nevertheless, despite these advancements, hallucination remains a significant challenge.

\noindent \textbf{Visual Hallucination in LVLMs} typically refers to cases where the generated text is inconsistent with the input image at the instance level~\cite{rohrbach2018object, zhou2023analyzing, li2023evaluating} or the faithfulness of the generated free-form answer~\cite{jing2023faithscore}. There are various potential factors that can cause hallucinations, including modality gap, training data bias, error accumulation, etc.~\cite{zhou2023analyzing}. To evaluate LVLM hallucination levels, CHAIR~\cite{rohrbach2018object} proposes measuring object hallucination rates in output captions. The POPE benchmark~\cite{li2023evaluating} assesses object hallucinations using binary ``Yes/No'' questions. Additionally, MME ~\cite{mme} provides a more challenging dataset for hallucination evaluation, which encompasses various hallucination types, such as object, attribute, counting, etc. AMBER~\cite{wang2023amber} supports both generative and discriminative tasks, covering existence, attribute, and relation hallucinations. While prior works have primarily focused on natural photographic images, this study constructs a new dataset and comprehensive benchmark to investigate hallucination in LVLMs when processing stylized images.

% \textbf{Hallucination Mitigation.} Existing hallucination mitigation methods can be categorized into tuning-based and decoding-based methods. Tuning-based methods focus on curating dedicated datasets~\cite{wang2024mitigating, zhang2024reflective, yang2024nullu, chen2024ict, yue2024less} and applying alignment training~\citep {xie2024v, sun2023aligning, xing2024mitigating} to mitigate modality gaps between images and text. However, dataset annotation is time-consuming, and model training is computationally expensive. In turn, training-free decoding-based techniques offer a more efficient solution~\cite{huang2024opera, an2024agla}. Recently, contrastive decoding strategies~\cite{leng2024mitigating, chen2024halc, chuang2023dola, zhuang2024game, suo2025octopus} based on visual comparisons have been proposed to further improve hallucination mitigation performance. Deco~\cite{wang2024mllm} and Attention Lens~\cite{jiang2024devils} highlight the importance of selecting the appropriate preceding layers to bridge the knowledge gap between vision and language. Unlike previous approaches, our proposed method SAVER leverages richer visual information—especially critical for stylized images—to dynamically revise the final output and reduce hallucinations more effectively.
\noindent \textbf{Hallucination Mitigation.} Existing methods fall into tuning-based and decoding-based categories. Tuning-based approaches curate specialized datasets~\cite{wang2024mitigating, zhang2024reflective,jing2024fgaif} and apply alignment training~\cite{xie2024v, sun2023aligning, xing2024mitigating}, but are costly in annotation and computation. In contrast, training-free methods~\cite{yin2024woodpecker} and decoding-based techniques~\cite{huang2024opera, an2024agla} are more efficient. Recent contrastive decoding methods~\cite{leng2024mitigating, chen2024halc, chuang2023dola} further enhance performance by leveraging visual comparisons. Deco~\cite{wang2024mllm} and Attention Lens~\cite{jiang2024devils} emphasize layer selection to mitigate hallucination. Our method, \textbf{SAVER}, differs from these two approaches by dynamically selecting non-hallucinated tokens without the need to train a detector. And SAVER leverages richer visual signals during layer selection for final output revision, thereby achieving better hallucination mitigation performance.
% This approach is especially effective for stylized images, resulting in a greater reduction in hallucination.

% \section{Style Dataset and Benchmark}
% In this section, we start by introducing our proposed style dataset, including the motivation, dataset construction and image generation procedure. We then describe our benchmark. We describe the full procedures. The dataset style images examples are shown in Fig. \ref{fig:all-images}.

\begin{figure}[t]
\begin{center}
\includegraphics[width=1\linewidth]{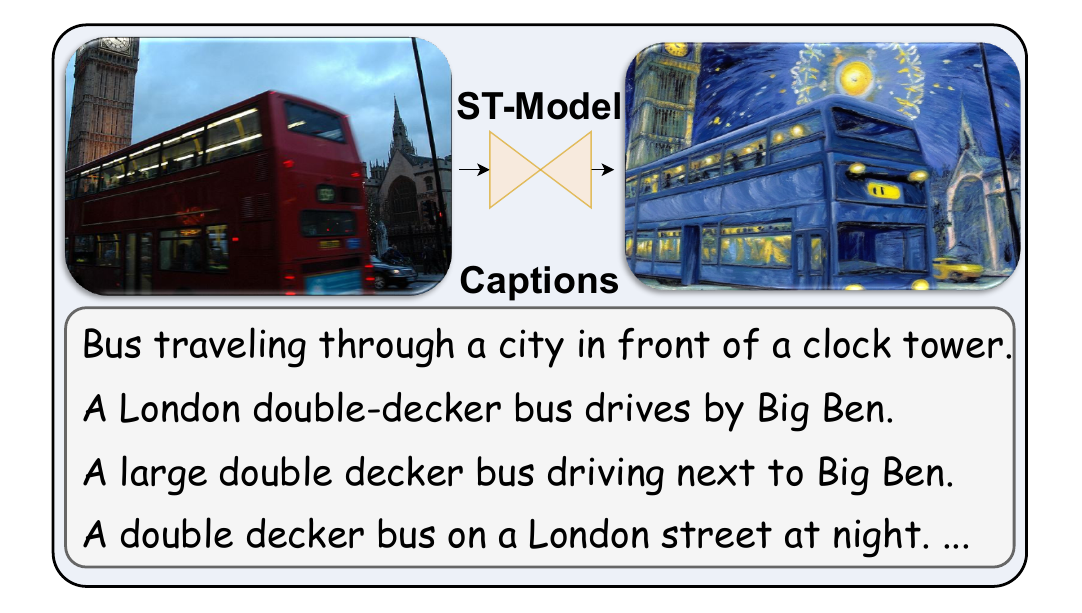}
\end{center}
    \caption{Top left: original image; top right: stylized image generated by Style Transfer (ST) model;  bottom: COCO captions listing all salient objects. 
    % The caption stays semantically faithful despite drastic color and texture changes, making this triplet a diagnostic test for LVLM hallucination under style shifts.
    }
    \label{fig:all-images}

\end{figure}
\section{Benchmarking Stylized Image Hallucination}

In this section, we introduce our proposed style dataset by outlining its motivation, construction process, and image generation pipeline. We then present the benchmark along with a detailed description of the procedures. An example style image from the dataset is shown in Fig.~\ref{fig:all-images}.

\label{Dataset}
% \textbf{Motivation.} We propose this dataset to explore the object hallucination problem in cross-style scenarios for LVLMs, assessing their ability to analyze images across diverse styles and handling inputs from less common domains. In this section, we detail the construction of the dataset and benchmark covering these varied styles, and introduce our hallucination mitigation method tailored for styled images.
% %
% Existing benchmarks for LVLMs predominantly evaluate models using photographic images, which we refer to as the ``Original'' style, reflecting their frequent use in applications. However, images span a broad range of styles, including Cartoon, Sketch, and others. Fig.~\ref{fig:all-images} illustrates a visual comparison between ``Original'' style images and various artistic styles. Objects in these alternative styles often exhibit distinct visual characteristics; for example, a cat with an unusual color rarely appears in ``Original'' style images. While humans can easily recognize such objects despite stylistic variations, it remains unclear whether LVLMs exhibit similar robustness. This gap may be attributed to insufficient exposure to diverse styles during training. To address this challenge, we introduce an evaluation framework to assess hallucination behavior in LVLMs when analyzing images across a spectrum of artistic styles.
\noindent\textbf{Motivation.} We propose this dataset to examine object hallucination in cross-style scenarios, evaluating LVLMs’ ability to analyze images from diverse domains. This section describes the construction of our style-diverse dataset and benchmark.
Existing LVLM benchmarks mainly use photographic (``Original'') images, reflecting their dominance in real-world applications. However, images also come in styles like Cartoon and Sketch, where objects may appear visually distinct (e.g., a cat with unusual colors). As shown in Fig.~\ref{fig:all-images}, such variations pose challenges for LVLMs, which may lack exposure to these styles during training. To address this, we introduce a framework to assess hallucination behaviors across a range of artistic styles.

\noindent\textbf{Dataset.} Follow previous hallucination works~\cite{rohrbach2018object, li2023evaluating}, we sampled images from the COCO dataset to construct a high-quality stylized dataset. Each selected image contains at least five annotated objects to create a challenging evaluation setting. We applied SOTA InstantStyle~\cite{wang2024instantstyle} model to generate style-transferred images in five styles: Cartoon, Game, Graffiti, Painting, and Sketch, ending up with 1,800 images. We carefully checked each pair of images and manually filter out the low-quality one, ensuring that each original image and its stylized versions share identical annotations, including object and caption labels.

\noindent\textbf{Benchmark.} We construct a benchmark using two metrics:
\begin{itemize}
    \item \textbf{CHAIR:} Caption Hallucination Assessment with Image Relevance (CHAIR)~\cite{rohrbach2018object} measures the proportion of hallucinated objects—those mentioned in the caption but absent in the image. We use two variants: CHAIRi (instance-level) and CHAIRs (sentence-level), defined as:
\begin{align}
\text{CHAIR}_i &= \frac{\{\text{hallucinated instances}\}}{\{\text{all mentioned instances}\}}, \nonumber \\
\text{CHAIR}_s &= \frac{\{\text{captions with hallucinations}\}}{\{\text{all captions}\}}.
\end{align}

    \item \textbf{POPE:} Polling-based Object Probing Evaluation (POPE)~\cite{li2023evaluating} evaluates hallucination by asking LVLMs ``Yes/No'' questions about object presence, using three negative sampling strategies: random, popular (top-$k$ frequent absent objects), and adversarial (top-$k$ absent objects ranked by co-occurrence with ground-truth). We follow the original POPE settings and generate 6 questions for each image, resulting in 1800 $\times$ 6 $\times$ 3 question image pairs.
\end{itemize}

\section{Style Aware Visual Early Revision}
\label{method}

\begin{figure*}[t]
\begin{center}
\includegraphics[width=1\linewidth]{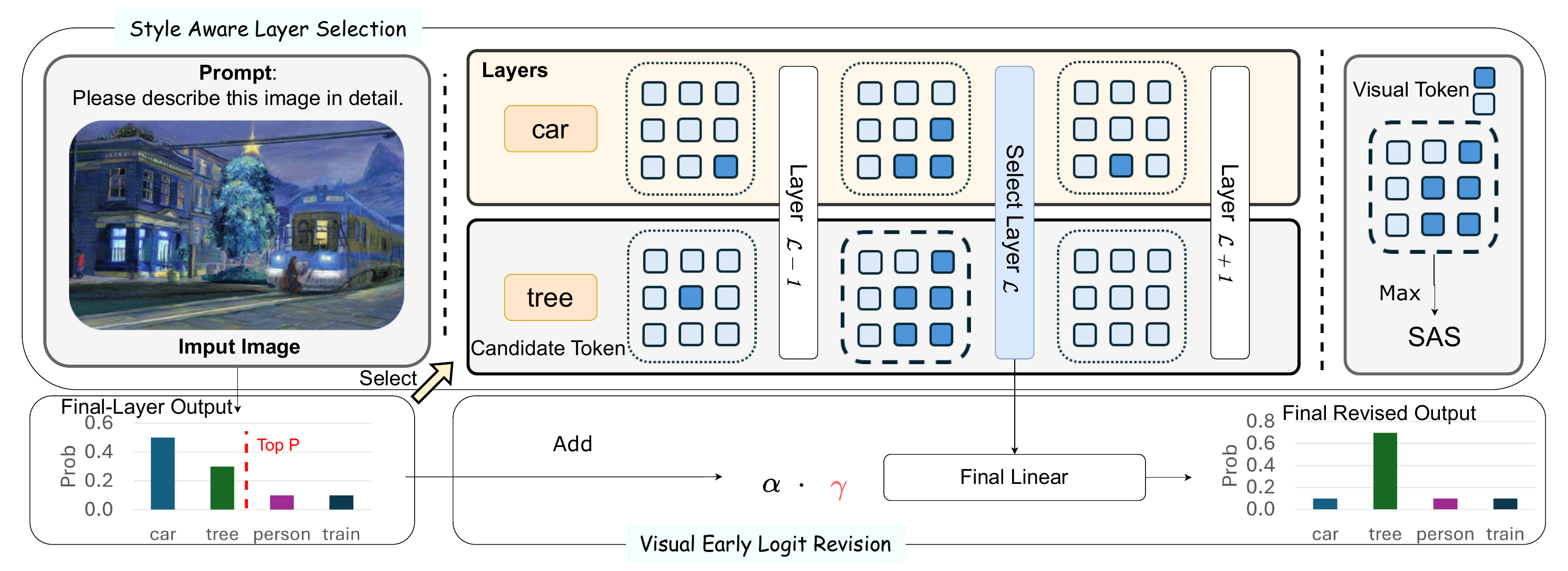}
\end{center}
   \caption{SAVER first selects the top-$p$ tokens, then chooses a layer via the Style-Aware Score, and finally revises the final-layer output. Darker visual tokens indicate higher confidence.}
\label{fig:method}
\end{figure*}

We conduct a comprehensive benchmark and study to uncover the underlying mechanisms that lead to object hallucination when processing stylized images. Guided by these findings, we propose \textbf{S}tyle-\textbf{A}ware \textbf{V}isual \textbf{E}arly \textbf{R}evision (\textbf{SAVER}), a novel inference-time strategy aimed at reducing hallucinations caused by style-transferred content. The overall design and workflow of SAVER are illustrated in Fig.~\ref{fig:method}.

\subsection{Preliminaries} 
LVLMs for text generation typically consist of three key components: a vision encoder, a projection module, and an autoregressive language model. The vision encoder first converts an input image into a sequence of visual tokens \( X_V = \{x_{v_1}, x_{v_2}, \dots, x_{v_P}\} \).  In parallel, a text prompt is tokenized into \( Q \) textual tokens \( X_C = \{x_{c_1}, x_{c_2}, \dots, x_{c_Q}\} \). Here \(P\) and \(Q\) are the lengths of the visual and textual tokens. The visual and textual embeddings are concatenated to form the model input \( X \), which is passed through an autoregressive language model composed of \( N \) stacked transformer decoder layers. At each layer \( i \), the model produces hidden states \( h^i = \{ h^i_0, h^i_1, \dots, h^i_{T-1} \} \), where \( T = P + Q \), \(P\) is the length of the visual tokens. During generation, the hidden state at the final position \( h^N_{T-1} \) is projected via an affine transformation \( \phi(\cdot) \), typically using an unembedding matrix \( W_U \in \mathbb{R}^{|V| \times d} \), to produce a logit distribution over the vocabulary \( V \).

% \usepackage{booktabs}
% \usepackage{multirow}
% \usepackage{graphicx}

\subsection{Stylization Amplifies Object Hallucination} 
We investigate the impact of image stylization on object hallucination in LVLMs, specifically in the context of image captioning. Our benchmark includes various LVLMs, covering both open-source models (e.g., LLaVA~\cite{liu2023visual}, MiniGPT-4~\cite{zhu2023minigpt}, InstructBLIP~\cite{instructblip}, TinyLLaVA~\cite{zhou2024tinyllava}, Phi-3-V~\cite{abdin2024phi}, Fuyu~\cite{fuyu-8b}, Idefics2~\cite{laurenccon2024matters}, mPLUG-Owl2~\cite{ye2024mplug},Qwen-VL~\cite{Qwen-VL}) and closed-source platforms (e.g., GPT-4o~\cite{openai2023gpt4}, Gemini-1.5-Pro~\cite{gemini_llm}).

To elicit diverse responses, we employ two prompt templates: a detailed prompt (\textit{``Please describe this image in detail''}) and a concise prompt (\textit{``Provide a one-sentence caption for the provided image.''}). The generated captions are evaluated using the CHAIR metric, with results summarized in Tab.~\ref{tab:table_chair_prompt1} and Tab.~\ref{tab:table_chair_prompt0} (Appendix). In addition, we assess hallucination across three splits of the Style-POPE benchmark, with detailed results visualized in Fig.~\ref{fig:Pope Style Benchmark} in the Appendix. 

Our experiments reveal a consistent increase in object hallucination when models process stylized images, such as those rendered in Cartoon, Game, Graffiti, Painting, and Sketch styles, compared to original photographs. This performance degradation highlights the models' reduced grounding capabilities under stylistic shifts.

% Furthermore, our analysis across prompt types shows that longer, more detailed prompts generally induce higher hallucination rates than shorter ones.

\begin{table*}[t]\small
\renewcommand{\arraystretch}{0.3}
\centering

\setlength{\tabcolsep}{6pt} 
{%
\begin{tabular}{@{}c|cccccccccc|cc|cc@{}}
\toprule
\multirow{2}{*}[-0.05em]{Model}                     & \multicolumn{2}{c}{\textbf{Cartoon}} & \multicolumn{2}{c}{\textbf{Game}} & \multicolumn{2}{c}{\textbf{Graffiti}} & \multicolumn{2}{c}{\textbf{Painting}} & \multicolumn{2}{c|}{\textbf{Sketch}} & \multicolumn{2}{c|}{\textbf{Original}} & \multicolumn{2}{c}{\textbf{Average}} \\ \cmidrule(l){2-15} 
                                           & Ci                & Cs               & Ci              & Cs              & Ci                & Cs                & Ci                & Cs                & Ci               & Cs               & Ci                & Cs                 & Ci               & Cs                \\ \midrule
GPT4o              & {10.3}     & 29.0             & {7.5}    & 23.0            & {7.9}      & 21.3              & {6.7}      & 20.0              & {5.6}     & 17.7             & {4.7}      & {16.7}      & {7.1}     & 21.3              \\
Gemini-1.5-PRO          & 13.5              & {25.3}    & 7.9             & {14.3}   & 13.1              & {18.7}     & 11.3              & {19.3}     & 7.0              & {13.3}    & 6.2               & 18.3               & 9.8              & {18.2}     \\ \midrule
LLaVA-1.5            & 12.0              & 43.3             & 9.1             & 31.7            & 11.4              & 37.7              & 11.2              & 34.7              & 11.4             & 37.7             & 6.8               & 26.7               & 10.3             & 35.3              \\
LLaVA-1.5-13b       & 10.7              & 38.7             & 10.3            & 35.0            & 11.0              & 37.3              & 9.6               & 33.3              & 10.0             & 31.7             & 7.4               & 27.3               & 9.8              & 33.9              \\
LLaVA-v1.6-mistral-7b & 10.9              & 25.7             & 8.8             & 22.0            & 9.9               & 24.0              & 10.4              & 26.7              & 10.1             & 25.7             & 6.0               & 19.3               & {9.4}     & 23.9              \\
InstructBLIP          & 13.7              & 45.0             & 9.4             & 36.7            & 12.0              & 37.3              & {8.9}      & 31.7              & 10.2             & 35.7             & 6.6               & 25.0               & 10.1             & 35.2              \\
MiniGPT-4           & 12.0              & 38.7             & 9.8             & 33.0            & 11.3              & 34.7              & 10.6              & 32.7              & 11.3             & 33.3             & 8.8               & 31.0               & 10.6             & 33.9              \\
Tinyllava        & 13.6              & 34.0             & 8.1             & {20.3}   & {9.3}      & {22.0}     & 10.4              & {24.0}     & {9.2}     & {22.3}    & 6.5               & {19.0}      & 9.5              & {23.6}     \\
Phi3V                & 11.3              & 27.3             & 9.8             & 25.7            & 11.3              & 20.7              & 9.6               & 26.7              & 9.8              & 25.7             & 6.8               & 22.3               & 9.8              & 24.7              \\
Fuyu                       & 17.7              & 60.3             & 17.1            & 60.7            & 24.7              & 61.3              & 17.7              & 55.0              & 19.8             & 57.7             & 10.8              & 48.7               & 18.0             & 57.3              \\
Idefics2-8b   & {9.5}      & {23.7}    & 8.6             & 23.3            & 9.9               & 24.7              & 11.4              & 31.7              & {9.2}     & 26.0             & 6.5               & 20.0               & 9.2              & 24.9              \\
mPLUG-Owl2            & 13.5              & 40.7             & 11.6            & 37              & 17.2              & 44.7              & 13.9              & 38.7              & 10.4             & 35               & 7.9               & 27.3               & 12.4             & 37.2              \\
Qwen-VL                    & 13.2              & 42.7             & {8.3}    & 30.3            & 14.8              & 39.3              & 10.2              & 30.7              & 10.2             & 35.0             & {5.0}      & 19.7               & 10.3             & 33.0              \\ \bottomrule
\end{tabular}%
\caption{Style-Chair Benchmark with prompt: \textit{``Please describe this image in detail''}. Lower indicate fewer hallucinations.}
\label{tab:table_chair_prompt1}
}

\end{table*}

\subsection{Visual Sensitivity with Style-Aware Score (SAS)}
% \label{sec:sas_definition}
To quantify the influence of visual style on token generation, we propose \textit{Style-Aware Score} (SAS), which measures the alignment between intermediate visual representations and the final output tokens. Specifically, SAS captures the contribution of early-layer visual embeddings to token prediction by analyzing the intermediate hidden states within the transformer decoder.
Given a set of candidate tokens selected from the final decoder layer’s logits, we compute the SAS by aggregating token logits across visual token positions and selected intermediate layers. Let \( \mathcal{L} \subset \{1, 2, \dots, N-1\} \) denote a predefined set of candidate transformer layers (e.g., early or deep layers). For each selected layers \( l \in \mathcal{L} \), we extract logits \( \mathbf{Z}_l \in \mathbb{R}^{T \times |V|} \) by projecting the hidden states \( h^l \) via the output embedding matrix \( W_U \), i.e., \( \mathbf{Z}_l = W_U h^l \). We then isolate the logits corresponding to the visual token positions, yielding \( \mathbf{Z}_l^{(v)} \in \mathbb{R}^{P \times |V|} \). The Style-Aware Score for a candidate token \( c \) at layer \( l \) is defined as:
\begin{equation}
    \text{SAS}_{l}(c) = \sum_{p=1}^{P} softmax (\mathbf{Z}_{l}^{(v)}( c))
\end{equation}
Where \( P \) is the visual token length. This score quantifies the model’s reliance on visual features during token generation. \textbf{A higher SAS implies stronger alignment between visual input and the generated token, indicating stronger grounding.} SAS provides a way to quantitatively assess a model’s sensitivity to visual inputs.

\subsection{Reduced Style Awareness Produces Hallucination}\label{sec:finding} 

Prior studies have shown that language model priors often dominate visual inputs, with attention skewed toward object-like tokens~\cite{wang2024mllm,jiang2024devils}. Motivated by these findings, we hypothesize that intermediate representations in early decoder layers encode varying degrees of visual awareness, which directly influences hallucination.
Fig.~\ref{fig:teaser} illustrates this hypothesis: tokens grounded in real visual objects (e.g., ``tree'') exhibit higher visual awareness, while hallucinated tokens show weaker spatial alignment. 
\begin{figure}[t]
  \centering
\includegraphics[width=1\linewidth]{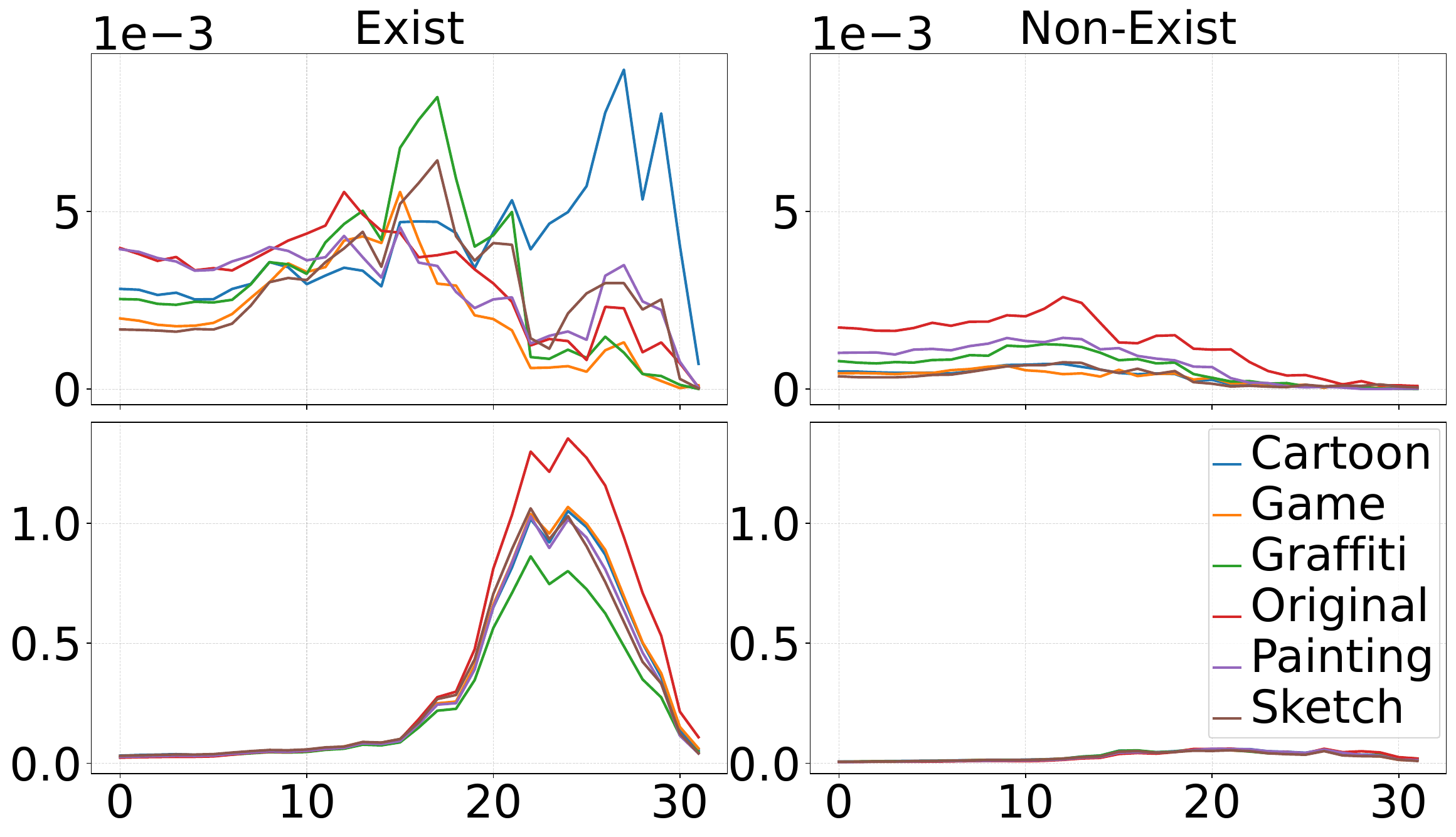}

  \caption{Style-awareness statistics across layer (Horizontal axis) using MiniGPT-4 (above) and InstructBlip.}
  \label{fig:finding2}

\end{figure}
To empirically validate this observation, we conduct a layer-wise analysis of SAS trajectories throughout the transformer stack using the prompt ``Please describe this image in detail". In the Style-POPE benchmark, each image is paired with six object existence questions (e.g.,``Is there a bottle in the image?''). A ``no'' label indicates the object is not present. For each question extract object names and their existence, and compute the average SAS values based on the predicted answer and the queried object.
As shown in Fig.~\ref{fig:finding2}, tokens corresponding to real objects consistently exhibit significantly higher SAS scores than those associated with non-existent (hallucinated) objects, especially in the early layers. We further observe that SAS distributions are style-dependent, with peaks often occurring in earlier layers of the model.
% To balance efficiency and performance, we adopt a unified layer setting for subsequent evaluations.

\begin{algorithm}[tb]

\caption{Style-Aware Visual Early Revision (SAVER)}
\label{alg:saver}
\textbf{Require} Input IDs \(\mathbf{x}\), beam width \(B\), early layers \(\mathcal{L}\), thresholds \(k, p\), scale \(\alpha\) \\
\textbf{Initialize beams} \(\{\mathbf{x}^{(b)}\}_{b=1}^{B}\)
\begin{algorithmic}[1] %[1] enables line numbers
\WHILE{condition}
\STATE \textbf{Decode:} Forward decoder to get final-layer logits \(\mathbf{z}^{N}_t\) and early-layer logits \(\{\mathbf{z}^{l}_t\}_{l \in \mathcal{L}}\)
\IF {first decoding step}
\STATE \textbf{Precompute SAS:} cache \(\{\operatorname{SAS}_l(c)\}_{l \in \mathcal{L},\ c \in V}\) \label{ln:start}
\ENDIF

\STATE \textbf{Candidate Filtering:} \(\mathcal{C}_t \leftarrow \textsc{TopK/TopP}(\mathbf{z}^{N}_t, k, p)\)
    \STATE \textbf{Style-Confidence Layer Selection:}\\
         \hspace{1.2em} \(\gamma = \max_{l \in \mathcal{L},\ c \in \mathcal{C}_t} \operatorname{SAS}_l(c)\)\\
         \hspace{1.2em} \(l^\star = \arg\max_{l \in \mathcal{L}} \max_{c \in \mathcal{C}_t} \operatorname{SAS}_l(c)\)
    \STATE \textbf{Logit Revision:} \(\hat{\mathbf{z}}_t = \mathbf{z}^{N}_t + \alpha \cdot \gamma \cdot (\mathbf{z}^{l^\star}_t \odot \mathbf{m}_t)\), where \(\mathbf{m}_t\) masks \(\mathcal{C}_t\)
    
    \STATE \textbf{Beam Update:} Update beams using revised logits \(\hat{\mathbf{z}}_t\)
\ENDWHILE
\STATE \textbf{return} best completed beam(s)
\end{algorithmic}
\end{algorithm}

% \usepackage{booktabs}
% \usepackage{multirow}
\begin{table*}[ht]\small
\centering
\setlength{\tabcolsep}{5.8pt} 
\renewcommand{\arraystretch}{0.2}

\begin{tabular}{@{}c|c|cccccccccccc|cc@{}}
\toprule
\multirow{2}{*}[-0.05em]{Model}        & \multirow{2}{*}[-0.05em]{Method} & \multicolumn{2}{c}{Cartoon}               & \multicolumn{2}{c}{Game}                 & \multicolumn{2}{c}{Graffiti}              & \multicolumn{2}{c}{Painting}             & \multicolumn{2}{c}{Sketch}                & \multicolumn{2}{c|}{Original}    & \multicolumn{2}{c}{Average}      \\ \cmidrule(l){3-16} 
                              &                         & Ci              & \multicolumn{1}{c|}{Cs} & Ci             & \multicolumn{1}{c|}{Cs} & Ci              & \multicolumn{1}{c|}{Cs} & Ci             & \multicolumn{1}{c|}{Cs} & Ci              & \multicolumn{1}{c|}{Cs} & Ci             & Cs              & Ci             & Cs              \\ \midrule
\multirow{7}{*}[-1.7em]{InstructBLIP} & Greedy                  & 13.7            & 45.0                    & 9.4            & 36.7                    & 12.0            & 37.3                    & 8.9            & 31.7                    & 10.2            & 35.7                    & 6.6            & 25.0            & 10.1           & 35.2            \\
                              & Beam                    & 11.8            & 41.3                    & 9.7            & 34.0                    & 10.8            & 34.7                    & \underline {7.8} & 27.3                    & 10.4            & 38.3                    & 7.1            & 28.7            & 9.6            & 34.1            \\
                              & Dola                    & 18.8            & 42.7                    & 19.3           & 48.3                    & 20.2            & 46.3                    & 17.8           & 43.7                    & 17.6            & 43.7                    & 12.7           & 39.0            & 17.7           & 44.0            \\
                              & OPERA                   & 12.4            & 38.3                    & 10.7           & 35.0                    & 12.2            & 37.3                    & 8.8            & 31.0                    & 9.9             & 34.7                    & 7.0            & 25.7            & 10.2           & 33.7            \\
                              & Deco                    & \underline {10.9} & \underline {36.3}         & \underline {8.9} & \textbf{28.0}           & \underline {10.6} & \underline {30.7}        & 9.4            & \underline {26.0}         & \underline {8.2}  & \textbf{25.3}           & \underline {5.5} & \underline {22.3} & \underline {8.9} & \underline {28.1} \\
                              & AGLA                    & 12.4            & 41.0                    & 9.6            & 36.0                    & 12.8            & 38.0                    & 9.9            & 33.0                    & 10.4            & 37.7                    & 7.0            & 27.3            & 10.4           & 35.5            \\
                              & SAVER(Ours)             & \textbf{9.9}    & \textbf{32.0}           & \textbf{8.0}   & \underline {30.0}         & \textbf{9.9}    & \textbf{28.0}           & \textbf{7.4}   & \textbf{25.3}           & \textbf{7.8}    & \underline {27.0}        & \textbf{5.3}   & \textbf{21.3}   & \textbf{8.1}   & \textbf{27.3}   \\ \midrule
\multirow{7}{*}[-1.7em]{LLaVA-1.5}    & Greedy                  & 12.0            & 43.3                    & 9.1            & 31.7                    & \textbf{11.4}   & 37.7                    & 11.2           & 34.7                    & 11.4            & 37.7                    & 6.8            & 26.7            & 10.3           & 35.3            \\
                              & Beam                    & 11.5            & 39.3                    & 9.3            & 32.3                    & \underline {11.9} & 33.3                    & 10.2           & 34.3                    & \underline {10.5} & 36.7                    & 6.0            & \underline {24.0} & 9.9            & 33.3            \\
                              & Dola                    & 15.4            & 44.0                    & 14.1           & 43.0                    & 16.9            & 49.0                    & 14.7           & 42.0                    & 14.9            & 46.7                    & 10.6           & 35.7            & 14.4           & 43.4            \\
                              & OPERA                   & 12.1            & 38.7                    & 9.6            & 33.0                    & 12.4            & 35.7                    & 10.1           & 31.7                    & 10.7            & \underline {31.7}         & \textbf{5.8}   & 24.7            & 10.1           & 32.6            \\
                              & Deco                    & \textbf{11.2}   & \textbf{31.3}           & \underline {8.9} & \underline {30.0 }        & \underline {11.9} & \underline {32.7}        & \underline {9.4} & \underline {31.3}         & \textbf{9.4}    & \textbf{30.7}           & \underline {5.9} & \textbf{21.3}   & \textbf{9.5}   & \textbf{29.6}   \\
                              & AGLA                    & \underline {11.3} & 38.0                    & 9.3            & 31.3                    & 12.5            & 40.0                    & 10.4           & 33.3                    & 11.9            & 39.3                    & 6.7            & 26.7            & 10.4           & 34.8            \\
                              & SAVER(Ours)             & \textbf{11.2}   & \underline {32.7}         & \textbf{8.8}   & \textbf{29.7}           & \textbf{11.4}   & \textbf{30.3}           & \textbf{9.3}   & \textbf{29.0}           & 10.6            & 32.0                    & 6.6            & 26.0            & \underline {9.7} & \underline {30.0} \\ \midrule
\multirow{7}{*}[-1.7em]{LLaVA-1.6}    & Greedy                  & 12.3            & \underline {27.0}                      & 8.8            & 22.0                    & \underline {9.9}             & 24.0                    & 10.4           & 26.7                    & 10.1            & 25.7                    & 6.0            & 19.3            & 9.4            & 23.9            \\
                              & Beam                    & 11.9            & 29.0                    & 9.6            & 24.3                    & 10.1            & 22.7                    & 10.7           & 26.7                    & 9.3             & 25.3                    & 5.9            & 18.0            & 9.6            & 24.3            \\
                              & Dola                    & 13.6            & 31.0                    & \underline {8.1}            & \underline {18.3}                    & 10.2            & \textbf{17.7}           & \underline {8.9}           & \textbf{18.7}           & \underline {8.7}             & \underline {20.0}                    & \textbf{5.1}   & \textbf{12.7}            & \underline {9.1}            & \textbf{19.7}   \\
                              & OPERA                   & \textbf{10.1}   & \textbf{23.0}           & 8.6            & 21.5         & 12.1            & 27.1                    & 9.8            & 25.6                    & 9.2             & 24.7           & 6.0            & 17.4            & 9.3            & 23.2            \\
                              & Deco                    & 12.8            & 30.3                    & 9.7            & 24.3                    & 10.9            & 25.3                    & \textbf{8.4}   & 22.0                    & 9.3             & 26.7                    & 6.1            & 18.7            & 9.5            & 24.6            \\
                              & AGLA                    & \underline{10.8}           & 28.0                    & 9.2            & 23.3                    & \underline {9.9}             & 24.0                    & 10.5           & 27.3                    & 8.9             & 25.7                    & 6.0            & 20.3            & 9.2            & 24.8            \\
                              & SAVER(Ours)             & 13.0            & 31.7                    & \textbf{7.4}   & \textbf{16.3}           & \textbf{9.7}    & \underline {20.3}                    & 9.0            & \underline {19.7}                    & \textbf{8.3}    & \textbf{18.7}                   & \underline {5.2}            & \underline {13.0}   & \textbf{8.8}   & \underline {20.0}            \\ \midrule
\multirow{7}{*}[-1.7em]{MiniGPT-4}    & Greedy                  & 12.0            & 38.7                    & 9.8            & 33.0                    & 11.3            & 34.7                    & 10.6           & 32.7                    & 11.3            & \underline{33.3}                    & 8.8            & 31.0            & 10.6           & 33.9            \\
                              & Beam                    & 11.5            & \underline {34.3}         & 10.7           & 34.3                    & \underline {9.5}  & 30.0                    & 10.0           & 33.7                    & 11.2            & 34.3                    & 8.1            & 28.7            & 10.2           & 32.6            \\
                              & Dola                    & 16.4            & 40.3                    & 12.3           & 34.3                    & 15.6            & 36.7                    & 13.4           & 35.0                    & 13.5            & 36.0                    & 9.8            & 28.7            & 13.5           & 35.2            \\
                              & OPERA                   & 11.5            & 34.7                    & 10.9           & 34.7                    & 10.4            & 31.0                    & 9.6            & 31.0                    & 10.7            & 34.3                    & 8.2            & 29.0            & 10.2           & 32.5            \\
                              & Deco                    & \underline {11.3} & 38.3                    & \underline {9.2} & \underline {29.3}         & 10.2            & \underline {28.3}         & \underline {8.6} & \underline {27.7}         & \underline {9.8}  & \textbf{28.0}           & \underline {7.0} & \underline {24.7} & \underline {9.4} & \underline {29.4} \\
                              & AGLA                    & 13.2            & 43.7                    & 11.2           & 43.2                    & 13.5            & 45.3                    & 11.5           & 39.0                    & 12.9            & 42.0                    & 10.7           & 42.3            & 12.2           & 42.6            \\
                              & SAVER(Ours)             & \textbf{9.4}    & \textbf{30.0}           & \textbf{8.8}   & \textbf{28.7}           & \textbf{9.3}    & \textbf{28.0}           & \textbf{8.5}   & \textbf{25.3}           & \textbf{8.1}    & \textbf{28.0}           & \textbf{5.8}   & \textbf{24.0}   & \textbf{8.3}   & \textbf{27.3}   \\ \bottomrule
\end{tabular}
\caption{CHAIR hallucination evaluation results (lower is better). Best values are bolded; second-best are underlined.}
\label{tab:my-table}
\end{table*}

\subsection{Our Method}
\label{sec:method}
As discussed in the previous section, object hallucination in stylized images arises from the model’s neglect of the corresponding visual patterns during decoding. To mitigate this issue, we propose \textbf{S}tyle-\textbf{A}ware \textbf{V}isual \textbf{E}arly \textbf{R}evision (\textbf{SAVER}), a test-time strategy that adjusts final predictions using early-layer representations. Notably, SAVER introduces no additional learnable parameters and can be seamlessly integrated into existing LVLMs and mitigate hallucination problems. 
SAVER operates during decoding and consists of three key steps: (1) identifying a candidate set of plausible tokens using top-$p$ filtering; (2) selecting the most visually grounded layer via the Style-Aware Score (SAS); and (3) revising the logits to better reflect visual evidence from that layer. The algorithm pipeline is presented in \textbf{Algorithm~\ref{alg:saver}}, which outlines the step-by-step implementation of SAVER at test time.

\noindent \textbf{Style-Aware Layer Selection.}  
Our previous analysis shows that correct tokens are highly correlated with the activation regions in early layers where visual features dominate. In this vein, we define a candidate set of transformer layers \( \mathcal{L} \subset \{1, \dots, N{-}1\} \) and identify a candidate token set \( \mathcal{C}_t \) by applying top-\(p\) filtering to the final-layer logits \( \mathbf{z}_t^N \in \mathbb{R}^{|V|} \) and top-\(k\) for the tokens. For each layer \( l \in \mathcal{L} \), we compute a style-confidence score based on the maximum Style-Aware Score (SAS; see Eq.~(2)) among the candidate tokens:
\begin{equation}
l^\star = \arg\max_{l \in \mathcal{L}} \sigma_l, \quad \gamma = \sigma_{l^\star}, \quad \text{where} \quad \sigma_l = \max_{c \in \mathcal{C}_t} \operatorname{SAS}_l(c).
\end{equation}
Here, \( \gamma \in [0,1] \) quantifies the influence of stylistic features at the selected layer \( l^\star \) and adaptively modulates visually relevant outputs. This mechanism allows SAVER to dynamically identify which layer provides the most relevant visual grounding at each decoding step, balancing between overfitting to style and ignoring visual context. The use of maximum SAS ensures that even if a single token strongly activates visual features at a certain layer, that signal is preserved in layer selection. In practice, we find that this adaptive grounding depth is critical for generalization across diverse styles and model architectures.

\noindent \textbf{Logit Revision.}  
SAVER refines the final-layer logits \( \mathbf{z}_t^N \) by incorporating evidence from the selected style-sensitive layer \( l^\star \). Let \( \mathbf{m}_t \in \{0,1\}^{|V|} \) denote a binary mask that activates only the candidate tokens in \( \mathcal{C}_t \), which suppresses noise and stabilizes the logit revision. The revised logits \( \hat{\mathbf{z}}_t \) are computed as:
\begin{equation}
\label{eq:revision}
\hat{\mathbf{z}}_t = \mathbf{z}_t^N + \alpha \cdot \gamma \cdot \left( \mathbf{z}_t^{l^\star} \odot \mathbf{m}_t \right),
\end{equation}
where \( \alpha \) is a scalar hyperparameter and \( \odot \) is element-wise multiplication. This formulation selectively amplifies predictions that are grounded in style-aware visual evidence, while suppressing tokens that may arise solely due to stylistic noise. By scaling with both \( \alpha \) and the layer-specific confidence \( \gamma \), the method modulates its correction strength based on how strongly the model attends to visual input at \( l^\star \). Since SAVER operates per decoding step, it enables fine-grained correction without modifying the backbone model or compromising generation fluency. This makes it applicable in real-world scenarios with diverse data distributions and effective even under domain shifts.

\section{Experiments}

\subsection{Implementation Details}
\label{implementation detail}

\textbf{Evaluation Models and Settings.}
We evaluate our proposed method and baseline approaches on four representative LVLMs:  InstructBLIP~\cite{instructblip}, MiniGPT-4~\cite{zhu2023minigpt}, and LLaVA-1.5~\cite{liu2023visual}, LLaVA-1.6~\cite{liu2024improved}. The maximum length of the generated sequence is set to 64 tokens, with a repetition penalty of 1.0. Unless otherwise specified, decoding is performed using a fixed temperature of 0 and top-$k{=}1$. For beam search, we adopt a beam width of 3 while maintaining the same temperature setting.

\noindent \textbf{Baselines.} We compare SAVER with two baseline decoding strategies (greedy decoding and beam search) as well as three SOTA hallucination mitigation methods, detailed as follows:
Dola~\cite{chuang2023dola} is specifically designed for alleviating hallucinations in factual tasks for LLMs by reducing shallow semantic influences to improve the factuality of the final layer’s output. OPERA~\cite{huang2024opera} dynamically penalizes overconfident tokens based on the emergence of aggregation patterns, while proposing a retrospective allocation strategy to avoid cases where hallucinations have already occurred. Deco~\cite{wang2024mllm} adaptively chooses relevant layers and integrates their knowledge into the final layer to adjust outputs. AGLA~\cite{an2024agla} uses an ensemble of global features for response generation and local features to mitigate hallucination. For all baseline methods, we use their official implementations and follow the recommended hyperparameter settings from the released source code to ensure fair comparisons.

\noindent \textbf{Benchmark and Metrics.}
We evaluate the effectiveness, generalizability, and captioning quality of our method across five challenging benchmarks in both stylized and real-world scenarios:
$\bullet$ \textbf{CHAIR.} Using the prompt  “\textit{Please describe the image in detail},” we assess hallucination rates with CHAIR$\mathrm{i}$ and CHAIR$\mathrm{s}$ on our constructed dataset. Additionally, we evaluate captioning quality using BLEU-1/2/3/4~\cite{papineni2002bleu}, METEOR~\cite{banerjee2005meteor}, and ROUGE-L~\cite{lin2004rouge} which can be found in Appendix.
$\bullet$ \textbf{POPE.} Following the official POPE protocol, we report F1 scores as the primary metric. In the Appendix, we further incorporate ACC, Precision, and Recall to comprehensively evaluate our method.
$\bullet$ \textbf{MME}~\cite{mme} is a practical benchmark encompassing 14 sub-tasks, including OCR, visual knowledge, object recognition, and relational reasoning.
$\bullet$ \textbf{Real-World Cases.} To evaluate performance beyond stylized images, we construct a dataset containing depth, thermal, medical, and RGB images with carefully designed query prompts. Additional details are available in the Appendix.
$\bullet$ \textbf{AMBER.}~\cite{wang2023amber} To evaluate hallucination in more dimisions such as attribute, relation, and existence, we conduct experiments on the AMBER benchmark. More details are provided in the Appendix.
\subsection{Experimental Results}
% \usepackage{booktabs}
% \usepackage{multirow}
% \usepackage{booktabs}
% \usepackage{multirow}
\begin{table}[t]\small
\renewcommand{\arraystretch}{0.3}
\centering

\setlength{\tabcolsep}{0.9pt} 
\begin{tabular}{@{}c|ccc|ccc|ccc|ccc@{}}
\toprule
\multirow{2}{*}[-0.2em]{Method} & \multicolumn{3}{c|}{Adversarial}          & \multicolumn{3}{c|}{Popular} & \multicolumn{3}{c|}{Random} & \multicolumn{3}{c}{Overall Avg} \\ \cmidrule(l){2-13} 
& M. & L. & I.  & M. & L. & I. & M.  & L. & I. & M.  & L. & Ins.          \\ \midrule
Dola   & 26.1   & 63.9  & 62.9   & 26.4  & 64.3  & 63.2   & 25.8   & 64.2   & 64.8 & 26.1   & 64.1    & 63.7  \\
Deco & \underline{67.0}  & \underline{73.3} & \textbf{73.6} & \underline{66.8} & \underline{74.6} & \underline{72.2} & \underline{70.8} & \underline{80.5} & \underline{82.7} & \underline{68.2}  & \underline{76.2}   & \underline{76.2} \\
Ours  & \textbf{67.6} & \textbf{75.3} & \underline{73.5}  & \textbf{67.4} & \textbf{77.6} & \textbf{72.7} & \textbf{73.6} & \textbf{84.4} & \textbf{84.2} & \textbf{69.5} & \textbf{79.1} & \textbf{76.8} \\ \bottomrule
\end{tabular}
\caption{Results on Style-POPE hallucination. ``M''., ``L.'' and ``I.'' stand for MiniGPT-4, LLaVA-1.5, and InstructBLIP.}
\label{average-pope}
\end{table}

\begin{table}[t]\small
\centering
\renewcommand{\arraystretch}{0.3}
\setlength{\tabcolsep}{1pt}

{%
\begin{tabular}{@{}c|c|cccc|c@{}}
\toprule
Model                         & Method & \begin{tabular}[c]{@{}c@{}}Scene\end{tabular} & \begin{tabular}[c]{@{}c@{}}Num.\end{tabular} & \begin{tabular}[c]{@{}c@{}}Text.\end{tabular} & \begin{tabular}[c]{@{}c@{}}Code.\end{tabular} & \begin{tabular}[c]{@{}c@{}}Total\\ Score\end{tabular} \\ \midrule
\multirow{3}{*}[-0.3em]{LLaVA-1.5} & Dola & 58.6 & 60.0  & \underline{82.5} & \textbf{57.5}      & \underline{258.6}   \\
                              & Deco   & \underline{85.7} & \underline{70.0}                                                            & 50.0                                                       & \underline{47.5}                                                     & 253.2                                                 \\
                              & Ours  & \textbf{88.6}                                                    & \textbf{77.5}                                                   & \textbf{115.0}                                             & 45.0                                                     & \textbf{326.1}                                        \\ \midrule
\multirow{3}{*}[-0.3em]{InstructBLIP} & Dola    & \underline{81.4}                                                             & \textbf{55.0}                                                   & \underline{57.5}                                                       & 45.0                                                     & 238.9                                                 \\
                              & Deco  & \underline{81.4}                                                             & 45.0                                                            & \textbf{65.0}                                              & \textbf{72.5}                                            & \underline{263.9}                                                 \\
                              & Ours  & \textbf{107.1}                                                   & \underline{50.0}                                                            & \underline{57.5}                                                       & \underline{55.0}                                                     & \textbf{269.6}                                        \\ \bottomrule
\end{tabular}%
}
\caption{Results on MME recognition related to the subtasks of commonsense reasoning, numerical calculation, text translation, and code reasoning.}
\label{tab:my-table_mme2}
\end{table}

\textbf{Experiments on Style-CHAIR.} 
\label{subsec.result}
As shown in Tab.~\ref{tab:my-table}, our method consistently reduces hallucinations across all styles, achieving the lowest average CHAIR across all evaluated models. Notably, SAVER achieves the best overall performance on MiniGPT-4 and InstructBLIP, clearly outperforming the SOTA method Deco. On LLaVA-1.5, SAVER achieves competitive results, closely matching the best scores. These results demonstrate the effectiveness of SAVER in mitigating object hallucinations for stylized images across diverse models and styles. Since captioning quality is critical for real-world applications, we further report the performance of different mitigation methods in terms of captioning quality in Tabs.~\ref{tab:my-table_blue_ins}, \ref{tab:my-table_blue_llava}, and~\ref{tab:my-table_blue_mini} in the Appendix. Compared to existing baselines, SAVER consistently exhibits outstanding text captioning performance.

\noindent \textbf{Experiments on Style-POPE and MME Benchmarks.}
As shown in Tab.~\ref{average-pope}, SAVER consistently outperforms prior methods under various configurations on the Style-POPE benchmark, demonstrating superior robustness in mitigating hallucinations across adversarial, popular, and random settings. The detailed results are shown in Tab.~\ref{tab:my-pope-mini}, Tab.~\ref{tab:my-pope-instruct}, and Tab.~\ref{tab:my-pope-llava} in the Appendix. Additionally, in Tab.~\ref{tab:my-table_mme2} and Tab.~\ref{tab:my-table_mme} (Appendix), SAVER achieves the highest scores across both perception and recognition tasks on the challenging MME benchmark, significantly improving performance for LLaVA-1.5 and InstructBLIP. Beyond stylized image captioning, SAVER exhibits strong generalizability to mitigate hallucinations in various practical challenging tasks. 

% offering improvements across a wide spectrum of visual understanding capabilities.

\begin{table}[t]\small
\renewcommand{\arraystretch}{0.3}
\centering
\begin{tabular}{@{}c|cc|cc|cc@{}}
\toprule
\multirow{2}{*}[-0.1em]{Method} & \multicolumn{2}{c|}{LLaVA-1.5} & \multicolumn{2}{c|}{MiniGPT-4} & \multicolumn{2}{c}{InstructBLIP} \\ \cmidrule(l){2-7} 
                        & ACC            & F1            & ACC            & F1            & ACC             & F1             \\ \midrule
Dola                    & \underline{62.2}           & 62.4          & \textbf{55.4}  & \underline{37.2}          & \textbf{65.8}   & 43.8           \\
Deco                    & 57.4           & \underline{69.4}          & 50.2           & \textbf{66.4} & 60.0            & \underline{68.2}           \\
Ours                    & \textbf{64.0}  & \textbf{71.8} & \underline{51.4}           & \textbf{66.4} & \underline{65.0}            & \textbf{68.6}  \\ \bottomrule
\end{tabular}
\caption{Average Results on Real-World Benchmarks.}
\label{average_real_world}
\end{table}

\noindent \textbf{Experiments on Real-World Benchmarks.} To further examine SAVER's practicability, we evaluate it on real-world scenarios, including depth, thermal, and medical images. As the average results shown in Tab.~\ref{average_real_world}, SAVER consistently achieves the highest performance in different modalities.
Specifically, it obtains the best average F1 scores of 71.8\% on LLaVA-1.5 and 68.6\% on InstructBLIP. These results further validate the effectiveness of SAVER in real-world scenarios. The detailed results and discussions can be found in the Appendix.
% \usepackage{booktabs}
% \usepackage{multirow}

\noindent \textbf{Experiments on AMBER Benchmarks.} Previous experiments comprehensively demonstrated SAVER's effectiveness in mitigating object hallucination. Herein, we further evaluate SAVER's generalizability on the AMBER dataset, which consists of three types of hallucinations, including existence, attribute, and relation. AMBER comprises 1,004 images, each paired with the corresponding designed questions and annotated labels. As shown in Tab.~\ref{tab:my-table_amber2}, SAVER consistently achieves the best or second-best Acc/F1 scores compared to the SOTA methods, exhibiting its strong generalizability to various hallucination types. This can be attributed to our visual early revision design that drives the model to pay more attention to visual signals.

% Tab.~\ref{tab:my-table_amber1} shows that 
% The results are presented in Tabs.~\ref{tab:my-table_amber2} and \ref{tab:my-table_amber1}. 
% Compared with previous methods, the reduction in both generative and discriminative hallucinations indicates that SAVER improves truthfulness with slightly Hal and Cog drop, while maintaining low levels of hallucination. Furthermore, SAVER reduces hallucinations in the existence, attribute, and relation categories, further demonstrating its significant improvement in truthfulness.

\begin{table}[t]
\centering
\renewcommand{\arraystretch}{0.3}

\begin{tabular}{@{}c|cc|cc|cc@{}}
\toprule
\multirow{2}{*}[-0.1em]{Method} & \multicolumn{2}{c|}{Existence}               & \multicolumn{2}{c|}{Attribute} & \multicolumn{2}{c}{Relation} \\ \cmidrule(l){2-7} 
& Acc & F1 & Acc & F1 & Acc & F1 \\ \midrule
Dola   & \underline{66.0}  & \underline{79.5} & 48.1 & 44.7  & 22.1 & 25.2 \\
Deco   & 63.1  & 62.5 & \underline{67.2} & \underline{53.5}  & \textbf{69.5} & \underline{48.1} \\
Ours  & \textbf{67.7} & \textbf{80.7} & \textbf{68.3} & \textbf{58.2} & \underline{69.1} & \textbf{62.5} \\ \bottomrule
\end{tabular}
\caption{LLaVA-1.5-7b results on AMBER.}
\label{tab:my-table_amber2}
\end{table}

% \begin{table}[t]\small
% \centering
% \renewcommand{\arraystretch}{0.6}
% \setlength{\tabcolsep}{1.5pt}
% \caption{Results on AMBER Generative and Discriminative tasks with LLaVA-1.5-7b.}
% \label{tab:my-table_amber1}
% \begin{tabular}{@{}c|cccc|cccc@{}}
% \toprule
% \multirow{2}{*}[-0.3em]{Method} & \multicolumn{4}{c|}{Generative} & \multicolumn{4}{c}{Descriminative} \\ \cmidrule(l){2-9} 
%                         & CHAIR$\downarrow$   & Cover$\uparrow$  & Hal$\downarrow$   & Cog$\downarrow$  & Acc    & P.    & R.   & F1    \\ \midrule
% Dola                    & 8.8     & 42.1   & \textbf{30.9}  & \textbf{2.2}  & 51.2  & 70.7    & 52.7     & 60.4 \\
% Deco                    & 8.8     & \textbf{47.1}   & 36.5  & 4.7  & 63.1   & 96.0     & 46.3     & 62.5  \\
% Ours             & \textbf{8.7}     & 46.3   & 33.9  & 3.5  & \textbf{69.1}   & 91.1   & 59.1     & \textbf{71.7}  \\ \bottomrule
% \end{tabular}
% \end{table}

% \usepackage{booktabs}
% \usepackage{multirow}

\subsection{Ablation Study}
Tabs.~\ref{tab:my-abl-1},~\ref{tab:my-abl-2},~\ref{tab:my-abl-3} and Fig.~\ref{fig:ablation} in the Appendix detail the ablation experimental results. We vary five components—scale factor $\alpha$, confidence threshold $p$, candidate set size $k$, number of image‑representative tokens $N_i$, and early‑exit depth—to quantify their contributions to hallucination mitigation and caption fluency. For scale factor, $\alpha=0.6$ consistently balances visual grounding and language quality, attaining the lowest hallucination scores on LLaVA‑1.5. For token filtering, higher thresholds reduce low confidence tokens. And $p=0.9$ achieves the strongest average results for MiniGPT‑4 and InstructBLIP, while lower $p$ preserves diversity in challenging styles at the cost of stability. In the candidate size, $k=20$ emerges as a robust optimum across models, and larger $k$ introduces spurious evidence. Visual awareness is best supported by moderate $N_i$ (50–100), as larger values can introduce low-confidence tokens that increase hallucination risks. Finally, the choice of early exit depth can greatly affect hallucinations. Experiments show that ``Standard'' provides the most stable performance.

\section{Conclusion}

% This work studies the high hallucination risk of existing LVLMs when understanding stylized images and how to mitigate it. By constructing a stylized dataset and comprehensive benchmark, we show that stylized inputs tend to cause higher hallucination rates. To understand the cause, we analyze correlations between generated and image tokens, revealing that late layers suppress visual content and emphasize language priors. To address this, we propose a training-free mitigation method, SAVER, which dynamically corrects generated tokens by retrieving optimal early layers with dense visual correlation. Extensive experiments demonstrate its effectiveness across various models, datasets, and tasks. We hope this work promotes the development of more trustworthy LVLMs and broadens their use in challenging scenarios.
This work studies the high hallucination risk of existing LVLMs when understanding stylized images and how to mitigate it. By constructing a stylized dataset and a comprehensive benchmark, we demonstrate that stylized inputs significantly increase hallucination rates. To explore the underlying causes, we analyze the correlation between generated tokens and image tokens, revealing that the later layers of LVLMs tend to suppress visual information and rely more heavily on language priors. To address this, we propose a training-free mitigation method, SAVER, which dynamically corrects generated tokens by retrieving optimal early layers with dense visual correlation. Extensive experiments across diverse models, datasets, and tasks validate the effectiveness of SAVER. We hope this work contributes to the development of more trustworthy LVLMs and facilitates their application in challenging scenarios.

\section{Acknowledgments}
This work was carried out at the Rapid-Rich Object Search (ROSE) Lab, School of Electrical \& Electronic Engineering, Nanyang Technological University (NTU), Singapore. This research is supported by the National Research Foundation, Singapore and Infocomm Media Development Authority under its Trust Tech Funding Initiative. Any opinions, findings and conclusions or recommendations expressed in this material are those of the author(s) and do not reflect the views of National Research Foundation, Singapore and Infocomm Media Development Authority.

\bibliography{aaai2026}
\onecolumn
\section*{Appendix}

\subsection{Style-POPE Benchmark}
\label{Pope benchmark}

\begin{figure*}[t]
\begin{center}
\includegraphics[width=0.9\linewidth]{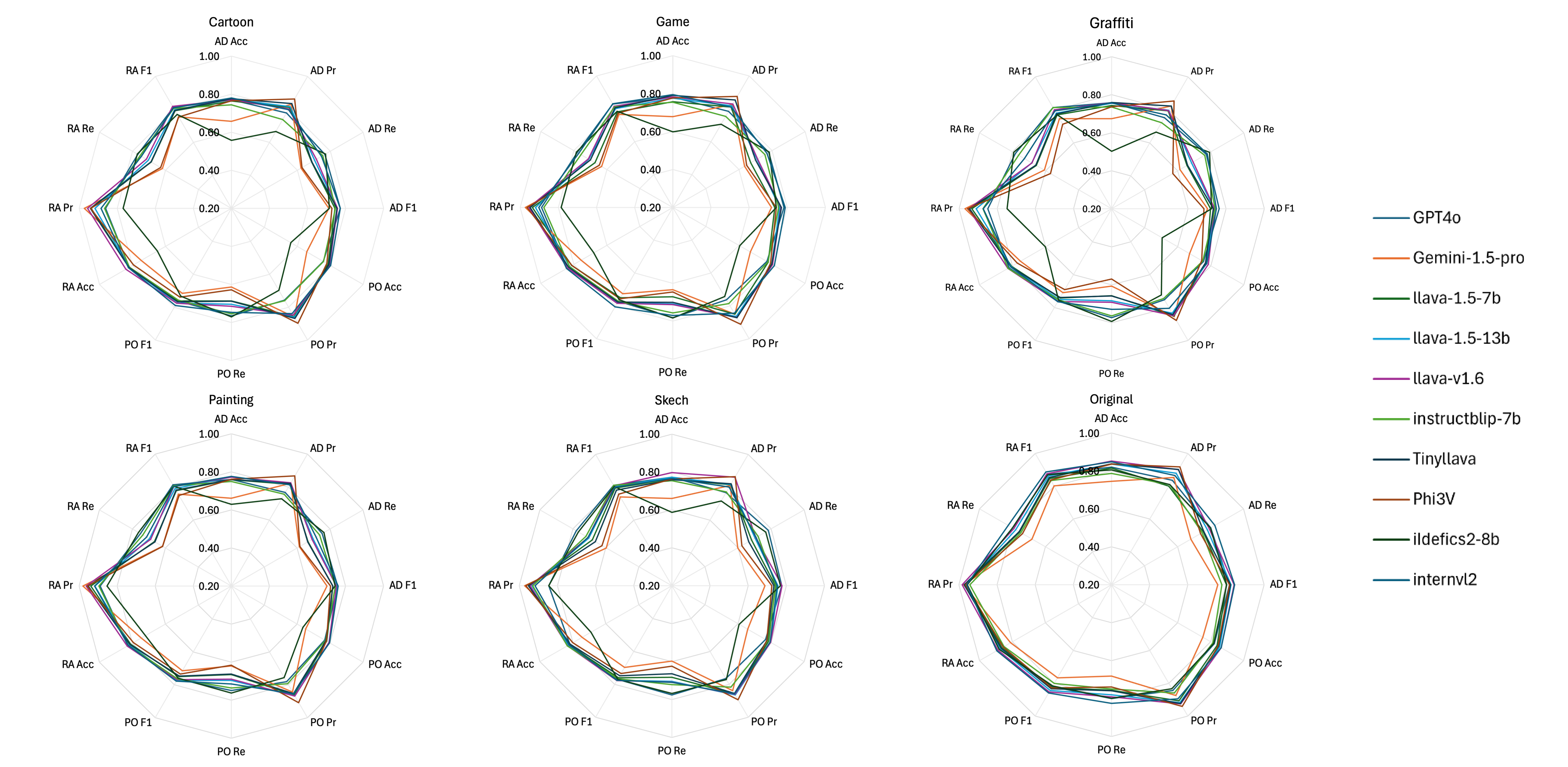}
\end{center}
   \caption{Style-POPE benchmark results.}
\label{fig:Pope Style Benchmark}
\end{figure*}

Here we present the Style-POPE hallucination benchmark results on existing LVLMs as shown in Fig. ~\ref{fig:Pope Style Benchmark}, including GPT-4o~\cite{openai2023gpt4}, Gemini-1.5-Pro~\cite{gemini_llm}, three variants of LLaVA (LLaVA-1.5-7B, LLaVA-1.5-13B, and LLaVA-1.6)~\cite{liu2023visual}, InstructBLIP-7B~\cite{instructblip}, TinyLLaVA~\cite{zhou2024tinyllava}, Phi3V~\cite{abdin2024phi}, IDEFICS2-8B~\cite{laurenccon2024matters}, and InternVL2~\cite{chen2024internvl}. We report accuracy, precision, recall, and F1 scores across Adversarial, Popular, and Random settings to comprehensively benchmark the LVLMs' hallucination performance in various styles. It can be observed that GPT-4o and Gemini-1.5-Pro consistently achieve superior hallucination evaluation scores, demonstrating their outstanding visual understanding performance. In addition, the evaluation results on original photographs largely outperform the other five styles across all POPE metrics, proving that LVLMs tend to produce more hallucinations with style images, aligning with our findings discussed in the main manuscript.

\begin{table*}[t]\small
\renewcommand{\arraystretch}{0.45}
\setlength{\tabcolsep}{6pt} 
\centering

{%
\begin{tabular}{@{}c|cccccccccc|cc|cc@{}}
\toprule
\multirow{2}{*}[-0.2em]{Model}                     & \multicolumn{2}{c}{\textbf{Cartoon}} & \multicolumn{2}{c}{\textbf{Game}} & \multicolumn{2}{c}{\textbf{Graffiti}} & \multicolumn{2}{c}{\textbf{Painting}} & \multicolumn{2}{c|}{\textbf{Sketch}} & \multicolumn{2}{c|}{\textbf{Original}} & \multicolumn{2}{c}{\textbf{Average}} \\ \cmidrule(l){2-15} 
                                           & Ci                & Cs               & Ci              & Cs              & Ci                & Cs                & Ci                & Cs                & Ci               & Cs               & Ci                 & Cs                & Ci                & Cs               \\ \midrule
GPT4o              & 6.2               & 7.0              & 3.1             & {3.7}    & 2.4               & {2.7}      & {4.3}      & {6.0}      & 3.6              & 5.0              & {0.7}       & {0.5}      & {3.4}      & {4.2}     \\
Gemini-1.5-PRO         & {4.9}      & {6.0}     & {3.0}    & 4.3             & {2.3}      & {2.7}      & 6.9               & 9.0               & {3.4}     & {4.7}     & 2.3                & 3.7               & 3.8               & 5.1              \\ \midrule
LLaVA-1.5          & 9.1               & 19.3             & 7.2             & 14.0            & 8.2               & 16.0              & 6.9               & 13.0              & 7.1              & 13.7             & 2.6                & 5.7               & 6.9               & 13.6             \\
LLaVA-1.5-13b       & 8.1               & 16.7             & 5.4             & 10.3            & 8.9               & 17.3              & 6.6               & 13.3              & 6.2              & 12.3             & 4.3                & 8.7               & 6.6               & 13.1             \\
LLaVA-v1.6-mistral-7b & 6.5               & 8.7              & {3.2}    & {4.3}    & {4.1}      & {5.3}      & 4.8               & 6.7               & 4.4              & {6.0}     & {1.2}       & {1.7}      & {4.0}      & {5.5}     \\
InstructBILP      & 5.8               & 10.3             & 5.0             & 9.0             & 5.2               & 8.3               & {3.0}      & {5.0}      & {4.3}     & 6.7              & 2.0                & 4.0               & 4.2               & 7.2              \\
MiniGPT-4-7b         & 11.4              & 27.0             & 9.1             & 20.3            & 10.5              & 21.0              & 9.0               & 21.7              & 9.4              & 23.3             & 5.4                & 14.0              & 9.1               & 21.2             \\
Tinyllava         & 6.1               & 12.3             & 5.6             & 11.7            & 5.8               & 12.3              & 5.2               & 11.3              & 6.6              & 14.0             & 2.5                & 5.3               & 5.3               & 11.2             \\
Phi3V                & 7.9               & 14.0             & 5.9             & 11.0            & 8.0               & 13.7              & 6.8               & 11.7              & 6.3              & 12.0             & 3.1                & 5.3               & 6.3               & 11.3             \\
Fuyu                       & 20.6              & 69.3             & 19.1            & 69.7            & 25.0              & 59.3              & 16.9              & 58.7              & 19.7             & 53.7             & 14.9               & 59.3              & 19.4              & 61.7             \\
Idefics2-8b   & 9.7               & 14.0             & 6.8             & 9.7             & 6.6               & 9.7               & 5.8               & 9.7               & 6.0              & 9.0              & 2.7                & 5.0               & 6.3               & 9.5              \\
mPLUG-Owl2                                & {5.6}      & {8.3}     & 5.6             & 8.3             & 6.2               & 8.0               & 4.7               & 6.0               & {4.3}     & 6.3              & 1.6                & 2.7               & 4.7               & 6.6              \\
Qwen-VL                                    & 8.4               & 15.0             & 6.4             & 13.0            & 9.5               & 14.7              & 7.1               & 12.0              & 7.4              & 14.3             & 3.1                & 6.3               & 7.0               & 12.6             \\ \bottomrule
\end{tabular}%
}
\caption{Style-Chair benchmark with prompt: \textit{``Provide a one-sentence caption for the provided image.''} Lower Ci and Cs indicate fewer hallucinations.}
\label{tab:table_chair_prompt0}
\end{table*}

\subsection{Style-Chair Benchmark}

\label{chair_benchmark}

% \usepackage{booktabs}
% \usepackage{multirow}
% \usepackage{graphicx}
\begin{table}[h]\small
\centering
{%
\begin{tabular}{@{}c|cccccccccccc|cc@{}}
\toprule
\multirow{2}{*}[-0.3em]{Method} & \multicolumn{2}{c}{Cartoon}             & \multicolumn{2}{c}{Game}               & \multicolumn{2}{c}{Graffiti}            & \multicolumn{2}{c}{Painting}           & \multicolumn{2}{c}{Sketch}             & \multicolumn{2}{c|}{Original} & \multicolumn{2}{c}{Average}   \\ \cmidrule(l){2-15} 
                        & Ci            & \multicolumn{1}{c|}{Cs} & Ci           & \multicolumn{1}{c|}{Cs} & Ci            & \multicolumn{1}{c|}{Cs} & Ci           & \multicolumn{1}{c|}{Cs} & Ci           & \multicolumn{1}{c|}{Cs} & Ci            & Cs            & Ci            & Cs            \\ \midrule
Deco                    & 14.0          & 46.0                    & 9.9          & \textbf{36.7}           & 17.0          & 35.7                    & 10.1         & 36.0                    & 10.0         & \textbf{33.7}           & \textbf{6.9}  & 37.0          & 11.3          & 37.5          \\
SAVER(ours)             & \textbf{11.8} & \textbf{37.7}           & \textbf{9.2} & \textbf{36.7}           & \textbf{13.5} & \textbf{34.7}           & \textbf{8.8} & \textbf{33.7}           & \textbf{8.2} & 34.3                    & 10.2          & \textbf{33.3} & \textbf{10.3} & \textbf{35.1} \\ \bottomrule
\end{tabular}%
}
\caption{CHAIR hallucination evaluation results with 512 max tokens. Lower scores indicate fewer hallucinations.}
\label{tab:my-table-512}
\end{table}

Here we present an extended analysis and results of our Chair benchmark using a different prompt.
Tab.~\ref{tab:table_chair_prompt0} summarizes hallucination rates on the Style-CHAIR benchmark using the prompt \textit{``Provide a one-sentence caption for the provided image.''} Similarly, the benchmark results show existing LVLMs have higher CHAIR scores in five styles. Compared to Tab.~\ref{tab:table_chair_prompt1}, the benchmark results in Tab.~\ref{tab:table_chair_prompt0} indicate that more detailed captions may produce more hallucinations, and LVLMs generate more hallucinations for style images. Moreover, variations in model architecture and training data also contribute to differences in hallucination rates.Tab.~\ref{tab:my-table-512} reports the Style-CHAIR benchmark results for LLaVA-1.5 with a maximum of 512 tokens. The average scores, 11.3/37.5 (Deco) vs. 10.3/35.1 (Ours), demonstrate the effectiveness of our method compared with the SOTA approach, and the per-style results follow the same trend as in the 64-token setting.

Additionally, we employ a SOTA style-transfer model and convert 50 stylized images across five styles back to photographic images to investigate whether the more severe hallucinations are caused by noise introduced by style transfer or by the style itself. As shown in Tab.~\ref{tab:my-table_multi_transfer}, the average scores (Ci/Cs: 14.9/46.0 vs. 14.3/34.0) indicate that stylized images indeed induce more hallucinations.
\begin{table}[h]\small
\centering
\begin{tabular}{@{}c|cccccccccc|cc@{}}
\toprule
\multirow{2}{*}[-0.2em]{\textbf{Image Type}} & \multicolumn{2}{c}{\textbf{Cartoon}}    & \multicolumn{2}{c}{\textbf{Game}}       & \multicolumn{2}{c}{\textbf{Graffiti}}  & \multicolumn{2}{c}{\textbf{Painting}}   & \multicolumn{2}{c|}{\textbf{Sketch}} & \multicolumn{2}{c}{\textbf{Average}} \\ \cmidrule(l){2-13} 
                                               & Ci            & \multicolumn{1}{c|}{Cs} & Ci            & \multicolumn{1}{c|}{Cs} & Ci           & \multicolumn{1}{c|}{Cs} & Ci            & \multicolumn{1}{c|}{Cs} & Ci                & Cs               & Ci                & Cs               \\ \midrule
Stylized                                            & 19.5          & 70.0                    & \textbf{9.1}  & \textbf{40.0}           & 10.6         & 40.0                    & \textbf{16.0} & 40.0                    & 19.1              & \textbf{40.0}    & \textbf{14.9}     & 46.0             \\
Photographic                                              & \textbf{13.7} & \textbf{50.0}           & \textbf{12.0} & \textbf{30.0}           & \textbf{9.3} & \textbf{20.0}           & \textbf{17.4} & \textbf{30.0}           & \textbf{19.0}     & \textbf{40.0}    & \textbf{14.3}     & \textbf{34.0}    \\ \bottomrule
\end{tabular}
\caption{CHAIR hallucination evaluation results for stylized images and photographic images.}
\label{tab:my-table_multi_transfer}
\end{table}
\subsection{Evaluation on Captioning Quality}

\label{sec.bleu}

While SAVER is explicitly designed to reduce object hallucinations, maintaining captioning fluency and informativeness is also crucial for practical applications. Tabs. \ref{tab:my-table_blue_ins}, \ref{tab:my-table_blue_llava}, and \ref{tab:my-table_blue_mini} present standard n–gram–based metrics (BLEU-1–4, METEOR, and ROUGE-L) evaluated on the generated captions of three models: InstructBLIP, LLaVA-1.5, and MiniGPT-4. 
Overall, these results demonstrate that SAVER can significantly mitigate hallucination while negligibly affecting captioning quality. 

% Across the five styles and original images, SAVER trails the strong beam-search baseline by no more than 1.5 BLEU-4 points and fewer than 1 METEOR point on average per model. This slight performance gap is an order of magnitude smaller than the substantial reductions in hallucination rates achieved by SAVER (see Sec. \ref{subsec.result}), clearly indicating that SAVER does not compromise caption quality in favor of brevity or generic content. Furthermore, SAVER consistently outperforms Deco and matches or surpasses AGLA, particularly where the latter exhibits weaknesses, consistently yielding the largest reduction in hallucinations. 

% Overall, these results demonstrate that SAVER can significantly mitigate the hallucination while negligibly affecting captioning quality, ensuring the practical usage of SAVER. 
% underscoring that accurate visual grounding and fluent language generation can indeed coexist.
\begin{table}[ht]
\centering

{%
\begin{tabular}{@{}c|c|cccccc@{}}
\toprule
Style                     & Method & Bleu1 & Bleu2 & Bleu3 & Bleu4 & METEOR & ROUGE\_L \\ \midrule
\multirow{6}{*}{Cartoon}  & Greedy & 24.4  & 11.7  & 6.4   & 3.5   & 17.2   & 20.4     \\
                          & Beam   & 27.2  & 16.5  & 10.1  & 6.2   & 20.0   & 26.4     \\
                          & OPERA  & 27.7  & 17.3  & 10.7  & 6.7   & 20.4   & 26.7     \\
                          & Deco   & 24.8  & 15.2  & 9.0   & 5.3   & 18.3   & 24.1     \\
                          & AGLA   & 26.2  & 16.9  & 10.4  & 6.4   & 20.0   & 25.6     \\
                          & Ours   & 25.1  & 15.9  & 9.7   & 6.0   & 19.4   & 24.6     \\ \midrule
\multirow{6}{*}{Game}     & Greedy & 24.4  & 11.8  & 6.6   & 3.6   & 17.4   & 20.4     \\
                          & Beam   & 27.3  & 16.9  & 10.3  & 6.3   & 20.7   & 26.0     \\
                          & OPERA  & 27.7  & 17.3  & 10.6  & 6.5   & 20.7   & 26.3     \\
                          & Deco   & 25.0  & 15.8  & 9.6   & 5.8   & 19.0   & 24.6     \\
                          & AGLA   & 26.7  & 17.4  & 10.8  & 6.7   & 20.6   & 26.0     \\
                          & Ours   & 25.0  & 16.2  & 10.0  & 6.1   & 19.8   & 24.5     \\ \midrule
\multirow{6}{*}{Graffiti} & Greedy & 24.2  & 11.0  & 5.6   & 2.8   & 16.6   & 20.7     \\
                          & Beam   & 27.2  & 16.4  & 9.8   & 5.9   & 19.7   & 26.1     \\
                          & OPERA  & 27.3  & 17.0  & 10.4  & 6.5   & 20.1   & 26.8     \\
                          & Deco   & 24.2  & 14.9  & 8.6   & 5.0   & 18.0   & 23.8     \\
                          & AGLA  & 26.4  & 17.0  & 10.3  & 6.2   & 19.5   & 25.8     \\
                          & Ours   & 24.5  & 15.5  & 9.3   & 5.7   & 18.8   & 24.3     \\ \midrule
\multirow{6}{*}{Painting} & Greedy & 24.2  & 11.2  & 6.0   & 3.2   & 17.0   & 20.0     \\
                          & Beam   & 27.3  & 16.8  & 10.1  & 6.1   & 20.5   & 25.7     \\
                          & OPERA  & 27.5  & 17.1  & 10.3  & 6.2   & 20.5   & 26.0     \\
                          & Deco   & 24.2  & 15.0  & 8.8   & 5.2   & 18.5   & 24.0     \\
                          & AGLA  & 26.1  & 16.7  & 10.3  & 6.5   & 20.2   & 25.6     \\
                          & Ours   & 24.6  & 15.7  & 9.3   & 5.5   & 19.3   & 24.1     \\ \midrule
\multirow{6}{*}{Sketch}   & Greedy & 24.0  & 11.3  & 6.1   & 3.4   & 16.9   & 19.8     \\
                          & Beam   & 26.9  & 16.7  & 10.2  & 6.4   & 20.4   & 25.5     \\
                          & OPERA  & 27.4  & 17.5  & 10.8  & 6.8   & 20.6   & 26.2     \\
                          & Deco   & 24.6  & 15.3  & 9.0   & 5.3   & 18.7   & 23.9     \\
                          & AGLA   & 26.4  & 17.2  & 10.5  & 6.4   & 20.3   & 25.2     \\
                          & Ours   & 25.0  & 16.1  & 9.8   & 6.0   & 19.7   & 24.6     \\ \midrule
\multirow{6}{*}{Original} & Greedy & 27.9  & 14.5  & 8.3   & 4.8   & 20.3   & 22.8     \\
                          & Beam   & 30.9  & 20.1  & 12.8  & 8.3   & 23.8   & 28.7     \\
                          & OPERA  & 31.5  & 21.0  & 13.6  & 8.8   & 24.1   & 29.5     \\
                          & Deco   & 28.3  & 18.8  & 11.9  & 7.5   & 22.2   & 27.2     \\
                          & AGLA   & 30.1  & 20.3  & 13.2  & 8.5   & 23.5   & 28.9     \\
                          & Ours   & 29.5  & 20.3  & 13.1  & 8.4   & 23.6   & 28.2     \\ \bottomrule
\end{tabular}%
}
\caption{Captioning quality scores on InstructBLIP.}

\label{tab:my-table_blue_ins}
\end{table}

\begin{table}[t]
\centering

{%
\begin{tabular}{@{}c|c|cccccc@{}}
\toprule
Style                     & Method & Bleu1 & Bleu2 & Bleu3 & Bleu4 & METEOR & ROUGE\_L \\ \midrule
\multirow{6}{*}{Cartoon}  & Greedy & 25.9  & 16.4  & 9.9   & 6.1   & 19.9   & 25.5     \\
                          & Beam   & 26.9  & 16.4  & 9.8   & 5.9   & 20.2   & 26.0     \\
                          & OPERA  & 26.9  & 16.7  & 10.1  & 6.2   & 20.2   & 26.1     \\
                          & Deco   & 25.5  & 15.3  & 9.0   & 5.4   & 18.7   & 24.4     \\
                          & AGLA~  & 26.2  & 16.1  & 9.6   & 5.8   & 19.7   & 25.4     \\
                          & Ours   & 25.3  & 14.8  & 8.4   & 4.9   & 18.5   & 24.0     \\ \midrule
\multirow{6}{*}{Game}     & Greedy & 26.8  & 17.2  & 10.5  & 6.4   & 20.3   & 25.7     \\
                          & Beam   & 26.9  & 16.7  & 10.1  & 6.1   & 20.4   & 25.8     \\
                          & OPERA  & 27.1  & 17.0  & 10.2  & 6.2   & 20.5   & 26.0     \\
                          & Deco   & 25.7  & 15.5  & 9.1   & 5.4   & 19.1   & 24.6     \\
                          & AGLA   & 26.5  & 16.8  & 10.1  & 6.2   & 20.3   & 25.5     \\
                          & Ours   & 25.2  & 15.3  & 8.8   & 5.0   & 18.5   & 24.1     \\ \midrule
\multirow{6}{*}{Graffiti} & Greedy & 26.0  & 16.7  & 10.4  & 6.6   & 20.0   & 25.3     \\
                          & Beam   & 26.5  & 16.5  & 10.1  & 6.3   & 20.1   & 25.6     \\
                          & OPERA  & 26.2  & 16.4  & 10.1  & 6.3   & 19.8   & 25.4     \\
                          & Deco   & 25.0  & 15.0  & 8.8   & 5.2   & 18.3   & 24.0     \\
                          & AGLA  & 25.8  & 16.3  & 9.9   & 6.1   & 19.7   & 25.0     \\
                          & Ours   & 24.7  & 14.6  & 8.3   & 4.8   & 18.2   & 23.7     \\ \midrule
\multirow{6}{*}{Painting} & Greedy & 26.6  & 17.0  & 10.3  & 6.2   & 20.3   & 25.8     \\
                          & Beam   & 26.7  & 16.2  & 9.7   & 5.9   & 20.1   & 25.7     \\
                          & OPERA  & 26.9  & 16.6  & 10.0  & 6.1   & 20.2   & 25.8     \\
                          & Deco   & 25.0  & 14.9  & 8.4   & 4.8   & 18.5   & 24.0     \\
                          & AGLA   & 26.5  & 16.4  & 9.7   & 5.7   & 20.2   & 25.5     \\
                          & Ours   & 25.2  & 14.8  & 8.3   & 4.8   & 18.5   & 23.9     \\ \midrule
\multirow{6}{*}{Sketch}   & Greedy & 25.7  & 16.5  & 10.1  & 6.1   & 20.1   & 24.7     \\
                          & Beam   & 26.5  & 16.6  & 10.1  & 6.2   & 20.1   & 25.4     \\
                          & OPERA  & 26.0  & 16.2  & 9.7   & 5.9   & 20.0   & 25.4     \\
                          & Deco   & 25.3  & 15.2  & 8.8   & 5.1   & 18.8   & 24.1     \\
                          & AGLA  & 25.9  & 16.4  & 9.9   & 5.9   & 19.8   & 24.8     \\
                          & Ours   & 24.9  & 14.8  & 8.3   & 4.7   & 18.3   & 23.9     \\ \midrule
\multirow{6}{*}{Original} & Greedy & 29.1  & 19.3  & 12.3  & 7.8   & 23.0   & 27.8     \\
                          & Beam   & 30.1  & 19.8  & 12.6  & 8.1   & 23.5   & 28.6     \\
                          & OPERA  & 30.1  & 19.8  & 12.7  & 8.2   & 23.3   & 28.5     \\
                          & Deco  & 27.9  & 17.7  & 10.8  & 6.5   & 21.5   & 26.1     \\
                          & AGLA  & 29.5  & 19.3  & 12.2  & 7.7   & 23.1   & 27.7     \\
                          & Ours   & 27.3  & 17.1  & 10.4  & 6.3   & 20.9   & 25.8     \\ \bottomrule
\end{tabular}%
}
\caption{Captioning quality scores on LLaVA-1.5. }
\label{tab:my-table_blue_llava}
\end{table}

\begin{table}[t]
 
\centering

{%
\begin{tabular}{@{}c|c|cccccc@{}}
\toprule
Style                     & Method & Bleu1 & Bleu2 & Bleu3 & Bleu4 & METEOR & ROUGE\_L\\ \midrule
\multirow{6}{*}{Cartoon}  & Greedy & 25.8  & 16.6  & 10.3  & 6.5   & 19.6   & 25.1     \\
                          & Beam   & 26.6  & 16.9  & 10.6  & 6.8   & 19.8   & 25.5     \\
                          & OPERA & 26.9  & 17.2  & 10.9  & 6.9   & 19.8   & 26.1     \\
                          & Deco  & 25.3  & 15.6  & 9.3   & 5.4   & 18.6   & 23.9     \\
                          & AGLA   & 19.5  & 12.8  & 8.0   & 5.0   & 17.6   & 20.9     \\
                          & Ours   & 25.3  & 16.0  & 9.8   & 5.9   & 19.0   & 24.9     \\ \midrule
\multirow{6}{*}{Game}     & Greedy & 26.2  & 17.2  & 10.9  & 7.0   & 20.1   & 25.1     \\
                          & Beam   & 26.7  & 17.4  & 11.0  & 7.0   & 20.3   & 25.7     \\
                          & OPERA  & 27.3  & 17.8  & 11.3  & 7.2   & 20.6   & 26.4     \\
                          & Deco   & 25.9  & 16.3  & 9.9   & 6.0   & 19.0   & 24.6     \\
                          & AGLA   & 19.2  & 12.6  & 7.8   & 4.9   & 17.5   & 20.7     \\
                          & Ours   & 25.7  & 16.6  & 10.2  & 6.3   & 19.6   & 25.1     \\ \midrule
\multirow{6}{*}{Graffiti} & Greedy & 25.5  & 16.5  & 10.4  & 6.6   & 19.6   & 25.0     \\
                          & Beam   & 26.3  & 16.7  & 10.4  & 6.6   & 19.9   & 25.3     \\
                          & OPERA  & 26.5  & 17.1  & 10.8  & 7.0   & 19.9   & 25.7     \\
                          & Deco   & 25.2  & 15.7  & 9.5   & 5.9   & 18.5   & 24.5     \\
                          & AGLA  & 20.4  & 13.4  & 8.3   & 5.2   & 17.9   & 21.6     \\
                          & Ours   & 25.5  & 16.4  & 10.0  & 6.1   & 19.0   & 25.3     \\ \midrule
\multirow{6}{*}{Painting} & Greedy & 26.3  & 17.3  & 10.9  & 7.0   & 20.3   & 25.6     \\
                          & Beam   & 26.6  & 17.3  & 10.8  & 6.9   & 20.4   & 25.9     \\
                          & OPERA  & 26.7  & 17.5  & 11.0  & 7.1   & 20.5   & 26.3     \\
                          & Deco   & 25.5  & 16.0  & 9.8   & 5.9   & 19.4   & 24.3     \\
                          & AGLA   & 19.9  & 13.1  & 8.1   & 5.0   & 17.8   & 20.8     \\
                          & Ours   & 25.5  & 16.3  & 10.0  & 6.1   & 19.2   & 25.0     \\ \midrule
\multirow{6}{*}{Sketch}   & Greedy & 26.6  & 17.5  & 11.1  & 7.1   & 20.4   & 25.4     \\
                          & Beam   & 27.2  & 17.9  & 11.5  & 7.4   & 21.1   & 26.2     \\
                          & OPERA  & 27.2  & 18.0  & 11.6  & 7.5   & 20.9   & 26.3     \\
                          & Deco  & 25.9  & 16.2  & 10.0  & 6.1   & 19.4   & 24.4     \\
                          & AGLA   & 20.3  & 13.4  & 8.3   & 5.1   & 18.1   & 21.2     \\
                          & Ours  & 26.0  & 16.7  & 10.2  & 6.3   & 19.9   & 25.2     \\ \midrule
\multirow{6}{*}{Original} & Greedy & 29.0  & 19.9  & 13.2  & 8.6   & 23.1   & 27.5     \\
                          & Beam   & 29.8  & 20.3  & 13.4  & 8.7   & 23.2   & 28.1     \\
                          & OPERA  & 30.2  & 20.9  & 13.9  & 9.0   & 23.7   & 28.8     \\
                          & Deco   & 28.5  & 18.9  & 12.1  & 7.7   & 22.1   & 26.5     \\
                          & AGLA   & 21.0  & 14.4  & 9.2   & 5.7   & 19.7   & 22.0     \\
                          & Ours   & 29.0  & 20.2  & 13.3  & 8.6   & 23.0   & 28.0     \\ \bottomrule
\end{tabular}%
}
\caption{Captioning quality scores on MiniGPT-4.}
\label{tab:my-table_blue_mini}
\end{table}

\subsection{Detailed Hallucination Mitigation Results on Style-POPE}

\label{Style-POPE}
Tabs.~\ref{tab:my-pope-mini}, \ref{tab:my-pope-instruct}, and \ref{tab:my-pope-llava} report the results on the Style-POPE benchmark for MiniGPT-4~\cite{zhu2023minigpt},  InstructBLIP~\cite{instructblip}, and LLaVA-1.5~\cite{liu2023visual} across three evaluation settings: Adversarial, Popular, and Random. Our proposed method, SAVER, consistently achieves superior hallucination mitigation performance across various experimental settings, evaluation metrics, and models. Specifically, SAVER enhances accuracy across every style–sampling combination for MiniGPT-4, with the most substantial improvement observed in the Painting style under Random sampling, where accuracy rises significantly from 60.2\% to 67.9\%. Similarly, for LLaVA-1.5 under the Random sampling setting of Original images, SAVER boosts accuracy from 82.8\% to 88.2\%, alongside proportional increases in precision and F1 scores. Furthermore, SAVER can still achieve performance gains on InstructBLIP. These results highlight SAVER’s significant improvements across diverse LVLM architectures and sampling strategies, underscoring its effectiveness in mitigating object hallucinations for style images.
\begin{table*}[ht]\small
\centering
\setlength{\tabcolsep}{5pt} 
\renewcommand{\arraystretch}{0.4}

{%
\begin{tabular}{@{}c|c|cccc|cccc|cccc@{}}
\toprule
\multirow{2}{*}[-0.3em]{Style}    & \multirow{2}{*}[-0.3em]{Method} & \multicolumn{4}{c|}{Adversarial}                       & \multicolumn{4}{c|}{Popular}                           & \multicolumn{4}{c}{Random}                             \\ \cmidrule(l){3-14} 
&                         & ACC             & Precision & Recall & \multicolumn{1}{c|}{F1}              & ACC             & Precision & Recall & \multicolumn{1}{c|}{F1}              & ACC             & Precision & Recall & F1              \\ \midrule
\multirow{3}{*}[-0.2em]{Cartoon}  & Dola                    & 52.10 & 57.30 & 16.60 & 25.70 & \textbf{53.40} & 63.10 & 16.60 & 26.20 & 48.70 & 46.40 & 16.60 & 24.40 \\
& Deco                   & 51.30 & 50.70 & 98.10 & 66.80 & 50.90 & 50.50 & 98.10 & 66.60 & 56.60 & 53.60 & 98.10 & 69.30 \\
& SAVER(Ours)                    & \textbf{53.80} & 52.10 & 94.20 & \textbf{67.10} & 53.10 & 51.70 & 94.20 & \textbf{66.70} & \textbf{63.90} & 58.60 & 94.20 & \textbf{72.30} \\ \midrule
\multirow{3}{*}[-0.2em]{Game}     & Dola                    & 53.60 & 60.90 & 19.90 & 30.00 & 54.80 & 66.10 & 19.90 & 30.60 & 51.70 & 54.70 & 19.90 & 29.20 \\
& Deco                  & 51.40 & 50.80 & 97.60 & 66.80 & 51.00 & 50.50 & 97.60 & 66.60 & 58.00 & 54.50 & 97.60 & 69.90 \\
& SAVER(Ours)                    & \textbf{54.60} & 52.60 & 93.80 & \textbf{67.40} & \textbf{54.20} & 52.30 & 93.80 & \textbf{67.20} & \textbf{64.40} & 59.10 & 93.80 & \textbf{72.50} \\ \midrule
\multirow{3}{*}[-0.2em]{Graffiti} & Dola                    & 50.60 & 53.20 & 10.20 & 17.10 & 51.60 & 59.00 & 10.20 & 17.40 & 49.40 & 47.20 & 10.20 & 16.80 \\
& Deco                 & 51.70 & 50.90 & 97.60 & 66.90 & 50.90 & 50.50 & 97.60 & 66.50 & 56.80 & 53.70 & 97.60 & 69.30 \\
& SAVER(Ours)                    & \textbf{54.20} & 52.40 & 93.70 & \textbf{67.20} & \textbf{53.60} & 52.00 & 93.70 & \textbf{66.90} & \textbf{63.40} & 58.30 & 93.70 & \textbf{71.90} \\ \midrule
\multirow{3}{*}[-0.2em]{Painting} & Dola                   & 49.70 & 48.90 & 12.80 & 20.30 & 50.20 & 50.90 & 12.80 & 20.40 & 48.70 & 45.30 & 12.80 & 19.90 \\
& Deco                 & 51.80 & 51.00 & 97.90 & 67.00 & 51.90 & 51.00 & 97.90 & 67.10 & 60.20 & 55.80 & 97.90 & 71.10 \\
& SAVER(Ours)                    & \textbf{55.40} & 53.10 & 92.80 & \textbf{67.50} & \textbf{54.70} & 52.60 & 92.80 & \textbf{67.20} & \textbf{67.90} & 61.90 & 92.80 & \textbf{74.30} \\ \midrule
\multirow{3}{*}[-0.2em]{Sketch}   & Dola                    & 14.10 & 51.70 & 56.90 & 13.70 & 12.00 & 53.20 & 65.40 & 13.70 & 10.40 & 51.50 & 56.20 & 13.70 \\
& Deco                & 52.20 & 51.20 & 98.10 & 67.30 & 51.90 & 51.00 & 98.10 & 67.10 & 58.90 & 55.00 & 98.10 & 70.50 \\
& SAVER(Ours)                    & \textbf{55.40} & 53.10 & 93.10 & \textbf{67.60} & \textbf{55.40} & 53.10 & 93.10 & \textbf{67.60} & \textbf{65.30} & 59.90 & 93.10 & \textbf{72.90} \\ \midrule
\multirow{3}{*}[-0.2em]{Original} & Dola                    & 54.20 & 55.20 & 45.10 & 49.60 & 55.00 & 56.20 & 45.10 & 50.10 & 55.90 & 57.50 & 45.10 & 50.60 \\
& Deco                   & 54.50 & 52.50 & 94.60 & 67.50 & 53.20 & 51.80 & 94.60 & 66.90 & 68.20 & 61.90 & 94.60 & 74.80 \\
& SAVER(Ours)                    & \textbf{60.40} & 56.80 & 86.90 & \textbf{68.70} & \textbf{60.20} & 56.60 & 86.90 & \textbf{68.60} & \textbf{75.00} & 70.20 & 86.90 & \textbf{77.70} \\ \midrule
\multirow{3}{*}[-0.2em]{Average}  & Dola                    & 45.72 & 54.53 & 26.92 & 26.07 & 46.17 & 58.08 & 28.33 & 26.40 & 44.13 & 50.43 & 26.80 & 25.77 \\
& Deco                   & 52.15 & 51.18 & 97.32 & 67.05 & 51.63 & 50.88 & 97.32 & 66.80 & 59.78 & 55.75 & 97.32 & 70.82 \\
& SAVER(Ours)                    & \textbf{55.63} & 53.35 & 92.42 & \textbf{67.58} & \textbf{55.20} & 53.05 & 92.42 & \textbf{67.37} & \textbf{66.65} & 61.33 & 92.42 & \textbf{73.60} \\ \bottomrule
\end{tabular}%
}
\caption{Style-POPE benchmark results on MiniGPT-4.}
\label{tab:my-pope-mini}
\end{table*}

\begin{table*}[ht]\small
\centering
\renewcommand{\arraystretch}{0.4}
\setlength{\tabcolsep}{5pt} 

{%
\begin{tabular}{@{}c|c|cccc|cccc|cccc@{}}
\toprule
\multirow{2}{*}[-0.3em]{Style}    & \multirow{2}{*}[-0.3em]{Method} & \multicolumn{4}{c|}{Adversarial}                       & \multicolumn{4}{c|}{Popular}                           & \multicolumn{4}{c}{Random}                             \\ \cmidrule(l){3-14}
                          &                         & ACC    & Precision & Recall & \multicolumn{1}{c|}{F1}     & ACC    & Precision & Recall & \multicolumn{1}{c|}{F1}     & ACC    & Precision & Recall & F1     \\ \midrule
\multirow{3}{*}[-0.2em]{Cartoon}  & Dola                     & \textbf{70.90} & 87.80 & 48.70 & 62.60          & \textbf{71.30} & 89.00 & 48.70 & 62.90          & 73.80 & 98.00 & 48.70 & 65.00          \\
                          & Deco                   & 66.90 & 61.80 & 88.40 & 72.80          & 63.30 & 58.90 & 88.40 & 70.70          & 78.90 & 74.30 & 88.40 & 80.70          \\
                          & SAVER(Ours)                    & 67.90 & 62.90 & 87.00 & \textbf{73.00} & 64.90 & 60.40 & 87.00 & \textbf{71.30} & \textbf{82.40} & 79.70 & 87.00 & \textbf{83.20} \\ \midrule
\multirow{3}{*}[-0.2em]{Game}     & Dola                   & \textbf{72.60} & 90.80 & 50.30 & 64.80          & \textbf{72.90} & 91.90 & 50.30 & 65.00          & 74.70 & 98.10 & 50.30 & 66.50          \\
                          & Deco                 & 67.40 & 62.40 & 87.90 & 73.00          & 64.20 & 59.60 & 87.90 & 71.00          & 79.50 & 75.30 & 87.90 & 81.10          \\
                          & SAVER(Ours)                    & 68.10 & 63.30 & 86.40 & \textbf{73.10} & 66.30 & 61.60 & 86.40 & \textbf{71.90} & \textbf{82.60} & 80.20 & 86.40 & \textbf{83.20} \\ \midrule
\multirow{3}{*}[-0.2em]{Graffiti} & Dola                    & \textbf{69.90} & 90.20 & 44.80 & 59.80          & 70.20 & 91.20 & 44.80 & 60.10          & 72.20 & 99.00 & 44.80 & 61.70          \\
                          & Deco                   & 66.30 & 61.20 & 89.20 & 72.60          & 64.30 & 59.50 & 89.20 & 71.40          & 79.20 & 74.30 & 89.20 & 81.10          \\
                          & SAVER(Ours)                    & 68.10 & 63.00 & 87.30 & \textbf{73.20} & \textbf{65.70} & 60.90 & 87.30 & \textbf{71.80} & \textbf{83.20} & 80.60 & 87.30 & \textbf{83.80} \\ \midrule
\multirow{3}{*}[-0.2em]{Painting} & Dola                    & \textbf{71.40} & 90.70 & 47.70 & 62.50          & \textbf{71.90} & 92.70 & 47.70 & 62.90          & 73.40 & 98.40 & 47.70 & 64.20          \\
                          & Deco                  & 68.30 & 63.20 & 87.90 & \textbf{73.50} & 65.20 & 60.50 & 87.90 & \textbf{71.60} & 82.60 & 79.50 & 87.90 & 83.50          \\
                          & SAVER(Ours)                    & 68.20 & 63.80 & 83.80 & 72.50          & 66.70 & 62.40 & 83.80 & 71.50          & \textbf{84.30} & 84.60 & 83.80 & \textbf{84.20} \\ \midrule
\multirow{3}{*}[-0.2em]{Sketch}   & Dola                    & 69.60 & 91.90 & 42.90 & 58.50          & 70.00 & 93.70 & 42.90 & 58.80          & 71.20 & 98.70 & 42.90 & 59.80          \\
                          & Deco                  & 69.90 & 65.10 & 86.00 & \textbf{74.10} & 68.10 & 63.30 & 86.00 & 72.90          & 83.20 & 81.50 & 86.00 & 83.70          \\
                          & SAVER(Ours)                    & \textbf{70.80} & 67.10 & 81.40 & 73.60          &\textbf{ 70.30}& 66.60 & 81.40 & \textbf{73.30} & \textbf{85.30} & 88.20 & 81.40 & \textbf{84.70} \\ \midrule
\multirow{3}{*}[-0.2em]{Original} & Dola                   & \textbf{75.20} & 90.40 & 56.30 & 69.40          & 75.70 & 91.80 & 56.30 & 69.80          & 77.90 & 99.20 & 56.30 & 71.90          \\
                          & Deco                   & 72.60 & 68.00 & 85.60 & \textbf{75.80} & 72.10 & 67.40 & 85.60 & 75.40          & 86.30 & 86.80 & 85.60 & 86.20          \\
                          & SAVER(Ours)                    & 73.60 & 69.80 & 83.00 & \textbf{75.80} & \textbf{74.10} & 70.50 & 83.00 & \textbf{76.20 }         & \textbf{86.90} & 90.10 & 83.00 & \textbf{86.40} \\ \midrule
\multirow{3}{*}[-0.2em]{Average}  & Dola                   & \textbf{71.60} & 90.30 & 48.45 & 62.93          & 72.00 & 91.72 & 48.45 & 63.25          & 73.87 & 98.57 & 48.45 & 64.85          \\
                          & Deco                   & 68.57 & 63.62 & 87.50 & \textbf{73.63} & 66.20 & 61.53 & 87.50 & 72.17          & 81.62 & 78.62 & 87.50 & 82.72          \\
                          & SAVER(Ours)                    & 69.45 & 64.98 & 84.82 & 73.53          & \textbf{68.00} & 63.73 & 84.82 & \textbf{72.67} & \textbf{84.12} & 83.90 & 84.82 & \textbf{84.25} \\ \bottomrule
\end{tabular}%
}
\caption{Style-POPE benchmark results on InstructBLIP.}
\label{tab:my-pope-instruct}
\end{table*}

\begin{table*}[ht]\small
\centering
\renewcommand{\arraystretch}{0.4}
\setlength{\tabcolsep}{5pt} 

{%
\begin{tabular}{@{}c|c|cccc|cccc|cccc@{}}
\toprule
\multirow{2}{*}[-0.3em]{Style}    & \multirow{2}{*}[-0.3em]{Method} & \multicolumn{4}{c|}{Adversarial}                       & \multicolumn{4}{c|}{Popular}                           & \multicolumn{4}{c}{Random}                             \\ \cmidrule(l){3-14} 
                          &                         & ACC    & Precision & Recall & F1     & ACC    & Precision & Recall & F1     & ACC    & Precision & Recall & F1     \\ \midrule
\multirow{3}{*}[-0.2em]{Cartoon}  & Dola                  & 67.60  & 67.30     & 68.30  & 67.80  & 67.60  & 67.30     & 68.30  & 67.80  & 67.90  & 67.80     & 68.30  & 68.10  \\
                          & Deco                    & 64.70  & 59.40     & 93.20  & 72.50  & 66.30  & 60.60     & 93.20  & 73.40  & 73.90  & 67.20     & 93.20  & 78.10  \\
                          & SAVER(Ours)                    & \textbf{70.10} & 64.80  & 88.10  & \textbf{74.70} & \textbf{73.10} & 67.70  & 88.10  & \textbf{76.60} & \textbf{81.70} & 78.10  & 88.10  & \textbf{82.80} \\ \midrule
\multirow{3}{*}[-0.2em]{Game}     & Dola                   & 66.80  & 66.80     & 66.90  & 66.90  & 68.20  & 68.60     & 66.90  & 67.80  & 66.30  & 66.10     & 66.90  & 66.50  \\
                          & Deco                  & 65.70  & 60.10     & 93.20  & 73.10  & 67.20  & 61.30     & 93.20  & 74.00  & 76.60  & 70.00     & 93.20  & 79.90  \\
                          & SAVER(Ours)                    & \textbf{70.20} & 65.00  & 87.70  & \textbf{74.60} & \textbf{74.20} & 69.00  & 87.70  & \textbf{77.20} & \textbf{82.90} & 80.10  & 87.70  & \textbf{83.70} \\ \midrule
\multirow{3}{*}[-0.2em]{Graffiti} & Dola                   & 63.90  & 64.10     & 63.10  & 63.60  & 64.30  & 64.60     & 63.10  & 63.90  & 62.20  & 61.90     & 63.10  & 62.50  \\
                          & Deco                  & 63.30  & 58.20     & 94.10  & 71.90  & 65.80  & 60.10     & 94.10  & 73.30  & 73.30  & 66.40     & 94.10  & 77.90  \\
                          & SAVER(Ours)                    & \textbf{69.30} & 64.20  & 87.40  & \textbf{74.00} & \textbf{72.50} & 67.30  & 87.40  & \textbf{76.10} & \textbf{81.80} & 78.60  & 87.40  & \textbf{82.80} \\ \midrule
\multirow{3}{*}[-0.2em]{Painting} & Dola                 & 66.40  & 66.00     & 67.80  & 66.90  & 67.60  & 67.50     & 67.80  & 67.60  & 67.10  & 66.90     & 67.80  & 67.30  \\
                          & Deco                   & 66.30  & 60.70     & 92.20  & 73.20  & 68.80  & 62.80     & 92.20  & 74.70  & 78.20  & 72.00     & 92.20  & 80.90  \\
                          & SAVER(Ours)                    & \textbf{70.80} & 66.00  & 85.60  & \textbf{74.50} & \textbf{74.80} & 70.40  & 85.60  & \textbf{77.20} & \textbf{83.30} & 81.90  & 85.60  & \textbf{83.70} \\ \midrule
\multirow{3}{*}[-0.2em]{Sketch}   & Dola                    & 65.10  & 61.90     & 63.50  & 47.50  & 65.10  & 61.90     & 63.50  & 47.50  & 63.70  & 61.90     & 62.80  & 48.60  \\
                          & Deco                  & 66.40  & 60.70     & 93.40  & 73.60  & 70.10  & 63.70     & 93.40  & 75.80  & 78.70  & 72.20     & 93.40  & 81.50  \\
                          & SAVER(Ours)                    & \textbf{72.10} & 66.80  & 87.70  & \textbf{75.80} & \textbf{75.80} & 70.90  & 87.70  & \textbf{78.40} & \textbf{84.20} & 81.90  & 87.70  & \textbf{84.70} \\ \midrule
\multirow{3}{*}[-0.2em]{Original} & Dola                   & 69.30  & 67.80     & 73.60  & 70.60  & 70.20  & 68.90     & 73.60  & 71.10  & 71.50  & 70.70     & 73.60  & 72.10  \\
                          & Deco                  & 68.40  & 61.70     & 97.00  & 75.50  & 70.30  & 63.20     & 97.00  & 76.50  & 82.80  & 75.60     & 97.00  & 85.00  \\
                          & SAVER(Ours)                    & \textbf{73.90} & 67.30  & 92.80  & \textbf{78.00} & \textbf{77.10} & 70.60  & 92.80  & \textbf{80.20} & \textbf{88.20} & 85.00  & 92.80  & \textbf{88.70} \\ \midrule
\multirow{3}{*}[-0.2em]{Average}  & Dola                    & 66.52  & 65.65     & 67.20  & 63.88  & 67.17  & 66.47     & 67.20  & 64.28  & 66.45  & 65.88     & 67.08  & 64.18  \\
                          & Deco                    & 65.80  & 60.13     & 93.85  & 73.30  & 68.08  & 61.95     & 93.85  & 74.62  & 77.25  & 70.57     & 93.85  & 80.55  \\
                          & SAVER(Ours)                    & \textbf{71.07} & 65.68  & 88.22  & \textbf{75.27} & \textbf{74.58} & 69.32  & 88.22  & \textbf{77.62} & \textbf{83.68} & 80.93  & 88.22  & \textbf{84.40} \\ \bottomrule
\end{tabular}%
}
\caption{Style-POPE benchmark results on LLaVA-1.5.}
\label{tab:my-pope-llava}
\end{table*}

\subsection{Hallucination Evaluation Results on MME dataset}

\label{MME}
MME is a more challenging dataset for LVLM hallucination evaluation, which comprises 14 fine-grained Visual Question Answering (VQA) tasks grouped into two distinct tracks: (i) ten perception-oriented tasks evaluating fundamental visual understanding abilities such as object existence, counting, and color identification; and (ii) four recognition-oriented tasks requiring advanced reasoning abilities, including commonsense inference, numerical calculations, text translation, and code understanding. 
% Each subtask contributes an absolute score, and track-level scores represent the aggregate of constituent subtasks, with higher total scores reflecting superior overall model capability.
As shown in the Tabs. \ref{tab:my-table_mme2} and \ref{tab:my-table_mme},  SAVER consistently achieves the highest scores across both perception and recognition tracks, regardless of the evaluated backbone architectures. These results indicate that our method not only effectively mitigates hallucinations but also enhances general visual comprehension. This improvement can be attributed to our method leveraging more visual information aligned with the text cues.
\begin{table*}[ht]
\centering
\setlength{\tabcolsep}{2.5pt} 

{%
\begin{tabular}{@{}c|c|cccccccccc|c@{}}
\toprule
Model                         & Method & Existence      & Count          & Position       & Color          & Posters        & Celebrity      & Scene          & Landmark       & Artwork        & OCR            & \begin{tabular}[c]{@{}c@{}}Total \\ Score\end{tabular} \\ \midrule
\multirow{3}{*}{LLaVA-1.5}        & Dola    & 170.0          & 55.0           & \textbf{126.7} & 108.3          & 79.3           & 69.1           & 117.8          & 114.0          & \textbf{76.0}  & 25.0           & 941.1                                                  \\
                              & Deco   & 165.0          & 93.3           & 91.7           & 95.0           & 138.8          & 103.5          & 151.5          & 112.0          & 58.3           & 85.0           & 1094.1                                                 \\
                              & SAVER(Ours)   & \textbf{185.0} & \textbf{113.3} & 106.7          & \textbf{125.0} & \textbf{144.9} & \textbf{117.4} & \textbf{155.3} & \textbf{134.8} & 66.0           & \textbf{117.5} & \textbf{1265.8}                                        \\ \midrule
\multirow{3}{*}{InstructBLIP} & Dola   & 165.0          & 68.3           & 50.0           & \textbf{148.3} & 60.2           & 60.2           & 110.0          & 50.0           & 87.8           & 87.5           & 876.8                                                  \\
                              & Deco   & \textbf{190.0} & 60.0           & 58.3           & 115.0          & 138.8          & 128.2          & 151.5          & \textbf{107.8} & 113.8          & \textbf{125.0} & 1188.3                                                 \\
                              & SAVER(Ours)   & \textbf{190.0} & \textbf{85.0}  & \textbf{66.7}  & 100.0          & \textbf{142.5} & \textbf{147.6} & \textbf{158.0} & 96.5           & \textbf{132.5} & 95.0           & \textbf{1213.8}                                        \\ \bottomrule
\end{tabular}%
}
\caption{Results on MME perception-related tasks. The best performance of each setting is bolded.}
\label{tab:my-table_mme}
\end{table*}
% Nonetheless, the relatively weaker performance observed on the code reasoning subtask identifies a clear avenue for future research: integrating SAVER with syntax-aware textual adapters could more effectively bridge visual representations with programmatic logic, thereby improving model performance in code-related tasks.

\subsection{Real-World Cases}

\label{sec.real-world}
\textbf{Benchmark Details.} 
To evaluate {SAVER} in more practical cases, we first collect a cross-modal dataset from three public datasets:
\begin{itemize}
\item \textbf{Depth–RGB (100 pairs).} Indoor scenes from NYU-Depth \cite{silberman2012indoor}; each dense depth map ($640\times480$) is aligned with a RGB photograph captured from the same viewpoint.
\item \textbf{Thermal–RGB (100 pairs).} Thermal and RGB frames from the KAIST Multispectral Pedestrian Benchmark \cite{hwang2015multispectral}; each long-wave infrared image ($640\times512$) has a pixel-wise registered RGB counterpart.
\item \textbf{Medical (50 images).} Single-channel X-rays from BenchLLM \cite{cai2024benchlmm}.
\end{itemize}

Following POPE, we formulate hallucination detection as a binary VQA task. For each image, we pose a concise, domain-specific ``Yes/No'' query that targets an object:
\begin{center}
\begin{tabular}{@{}cc@{}}
\toprule
\textbf{Modality} &  \textbf{Prompt (example)} \\
\midrule
Thermal & \textit{``Is there a person in the image?''} \\
Depth   & \textit{``Is there a laptop on the desk?''} \\
Medical & \textit{``Does the picture contain lungs?''} \\
\bottomrule
\end{tabular}
\end{center}

\begin{figure*}[h]
\begin{center}
\includegraphics[width=1\linewidth]{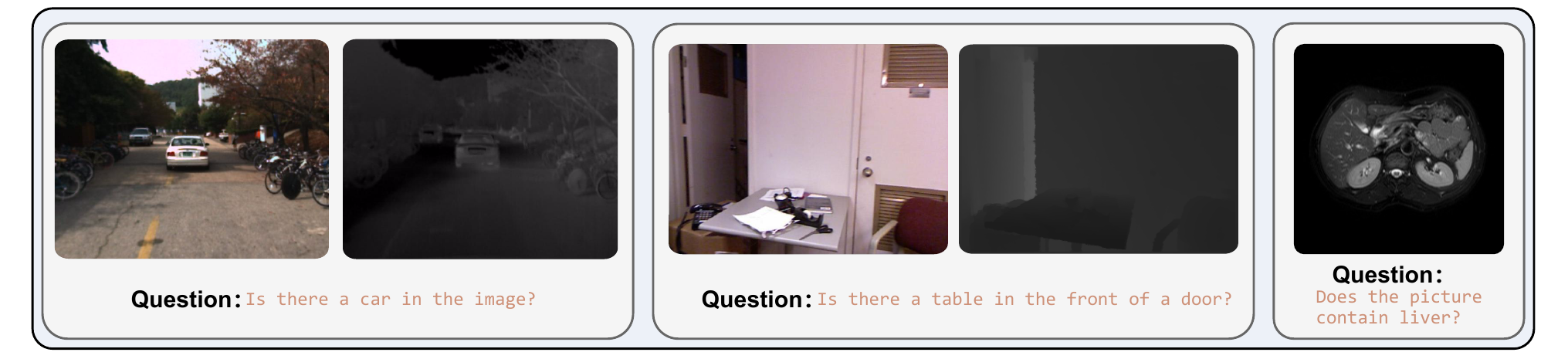}
\end{center}
   \caption{Example images of Real-World Cases. Left: Thermal-RGB images, middle: Depth-RGB images,   and Right: Medical images. }
\label{fig:realword_sample}
\end{figure*}

Objects are selected to (i) exhibit clear visual signatures when present, (ii) occur frequently in text-only corpora to align with real-world scenarios, and (iii) target the specific practical usage. This benchmark comprises 450 images paired with corresponding ``Yes/No'' queries, providing a rigorous test of LVLM robustness against hallucination. Some example images are shown in Fig. ~\ref{fig:realword_sample}.

\textbf{Experiments on Real-World Benchmark.}
Tab. \ref{tab:my-table_real} reports results on five real-world modalities—\emph{Depth}, \emph{Depth-RGB}, \emph{Thermal}, \emph{Thermal-RGB}, and \emph{Medical}—evaluated with three LVLMs: LLaVA-1.5 \cite{liu2023visual}, MiniGPT-4 \cite{zhu2023minigpt}, and InstructBLIP \cite{instructblip}. We compare our \textbf{SAVER} with two baseline hallucination mitigation approaches, Dola \cite{chuang2023dola} and Deco \cite{wang2024mllm}. SAVER exhibits the highest F1 scores under various experimental settings. These findings further demonstrate that our method can be effectively extended to challenging real-world data for hallucination mitigation.
\begin{table*}[t]\small
\centering
\setlength{\tabcolsep}{18pt}

{%
\begin{tabular}{@{}c|c|cc|cc|cc@{}}
\toprule
\multirow{2}{*}[-0.3em]{Style}       & \multirow{2}{*}[-0.3em]{Method} & \multicolumn{2}{c|}{LLaVA-1.5} & \multicolumn{2}{c|}{MiniGPT-4} & \multicolumn{2}{c}{InstructBLIP} \\ \cmidrule(l){3-8} 
                             &                         & ACC            & F1            & ACC            & F1            & ACC             & F1             \\ \midrule
\multirow{3}{*}{Depth}       & Dola                    & 51.0           & 47.3          & 54.0           & 23.3          & 50.0            & 0.0            \\
                             & Deco                    & 57.0           & 67.2          & 49.0           & 65.3          & 49.0            & 55.7           \\
                             & Ours            & 61.0           & 66.7          & 48.0           & 63.4          & 50.0            & 41.9           \\ \midrule
\multirow{3}{*}{Depth-RGB}   & Dola                    & 64.0           & 70.0          & 54.0           & 45.2          & 65.0            & 47.8           \\
                             & Deco                    & 57.0           & 69.9          & 52.0           & 67.6          & 70.0            & 72.7           \\
                             & Ours             & 63.0           & 72.6          & 51.0           & 66.2          & 82.0            & 82.4           \\ \midrule
\multirow{3}{*}{Thermal}     & Dola                    & 64.0           & 69.5          & 56.0           & 52.2          & 79.0            & 77.9           \\
                             & Deco                    & 65.0           & 73.7          & 50.0           & 66.7          & 59.0            & 70.5           \\
                             & Ours            & 71.0           & 76.4          & 50.0           & 66.7          & 64.0            & 73.1           \\ \midrule
\multirow{3}{*}{Thermal-RGB} & Dola                    & 76.0           & 80.0          & 63.0           & 65.4          & 85.0            & 86.0           \\
                             & Deco                    & 58.0           & 69.6          & 50.0           & 66.7          & 60.0            & 71.4           \\
                             & Ours             & 71.0           & 76.4          & 60.0           & 70.6          & 65.0            & 73.7           \\ \midrule
\multirow{3}{*}{Medical}     & Dola                    & 56.0           & 45.0          & 50.0           & 0.0           & 50.0            & 7.4            \\
                             & Deco                    & 50.0           & 66.7          & 50.0           & 65.8          & 62.0            & 70.8           \\
                             & Ours             & 54.0           & 66.7          & 48.0           & 64.9          & 64.0            & 71.9           \\ \midrule
\multirow{3}{*}{Average}     & Dola                    & 62.2           & 62.4          & \textbf{55.4}  & 37.2          & \textbf{65.8}   & 43.8           \\
                             & Deco                    & 57.4           & 69.4          & 50.2           & \textbf{66.4} & 60.0            & 68.2           \\
                             & Ours             & \textbf{64.0}  & \textbf{71.8} & 51.4           & \textbf{66.4} & 65.0            & \textbf{68.6}  \\ \bottomrule
\end{tabular}
}
\caption{Results on Depth, Thermal, and Medical images.}
\label{tab:my-table_real}
\end{table*}
\begin{figure*}[t]
\begin{center}
\includegraphics[width=0.9\linewidth]{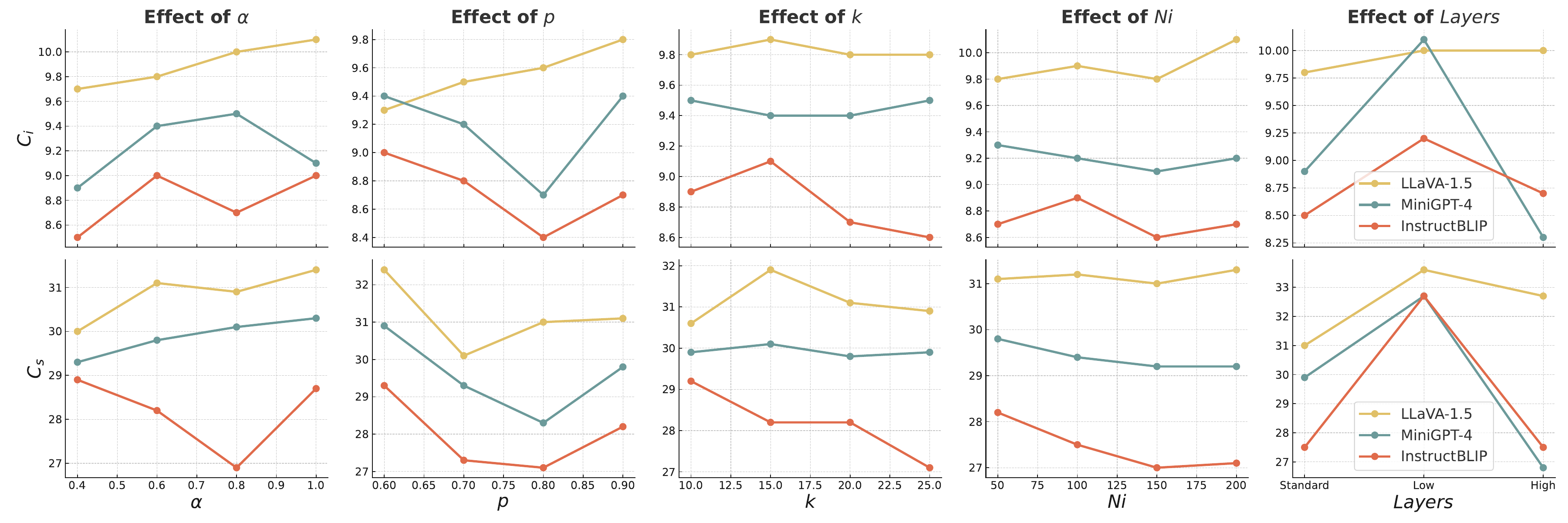}
\end{center}
   \caption{Ablation studies on hyperparameters.}
\label{fig:ablation}
\end{figure*}

\subsection{Ablation Study on Hyperparameters}
\label{app.ablation}

Here, we present the detailed ablation studies on five key hyperparameters. The average results are illustrated in Fig.~\ref{fig:ablation}, and The detailed results can be found in Tabs.~\ref{tab:my-abl-1},~\ref{tab:my-abl-2}, and~\ref{tab:my-abl-3}.

\noindent \textbf{Effect of the Scale Factor $\boldsymbol{\alpha}$.}
Tuning the scale factor $\alpha \in \{0.4, 0.6, 0.8, 1.0\}$ modulates the contribution of visual evidence in the fused representation. Higher values of $\alpha$ enhance visual guidance and generally reduce hallucinations; however, large values occasionally result in overly constrained generation.

\noindent \textbf{Effect of the Confidence Threshold $\boldsymbol{p}$.}
The confidence threshold~$p \in \{0.6, 0.7, 0.8, 0.9\}$ controls token selection by filtering candidate tokens based on confidence. Higher thresholds more aggressively suppress hallucinations but may  remove relevant content, adversely affecting fluency. Tab.~\ref{tab:my-abl-2}  demonstrates this trade-off, where lower thresholds ($p=0.6$) yield fewer hallucinations on stylized inputs, whereas intermediate to high thresholds ($p=0.8$–$0.9$) perform optimally on Original images. This pattern is most evident with MiniGPT-4 and InstructBLIP, suggesting mid-range~$p$ provides an optimal balance between robustness and fluency.

\noindent \textbf{Effect of the Candidate Set Size $\boldsymbol{k}$.} 
We also examine the effect of the number of candidate tokens~$k \in \{10, 15, 20, 25\}$. Low $k$ value excludes potentially informative tokens, hindering performance on rare styles, while excessively high $k$ introduces noisy tokens. From Tabs.~\ref{tab:my-abl-1}, \ref{tab:my-abl-2}, and \ref{tab:my-abl-3}, we consistently observe optimal performance at $k=20$, achieving the lowest average hallucination counts (e.g., LLaVA-1.5 in Tab.~\ref{tab:my-abl-3}).

\noindent \textbf{Effect of Candidate Image Representative Tokens $\boldsymbol{N_i}$.}  
We adjust the number of candidate image representations~$N_i \in \{50, 100, 150, 200\}$ to assess its influence. Moderate values ($N_i = 50$ or $100$) consistently provide the best trade-off by ensuring sufficient diversity to mitigate hallucinations during the decoding process (Tab.~\ref{tab:my-abl-2}, LLaVA-1.5). Higher values ($N_i \geq 150$) rarely yield further improvements.

\noindent \textbf{Interaction with Early-Exit Depth.}  
Following \cite{wang2024mllm}, we adopt the same layer configuration (20-29 layers) as the ``\textit{Standard}'' and investigate two alternative exit strategies: Extract 10 layers at equal intervals with in the early 20 layers (\textit{low}) and final 20 layers (\textit{high}), to evaluate the impact of depth. The ``\textit{Standard}'' policy consistently achieves the best overall hallucination mitigation results on stylized inputs. Notably, the higher layer exhibits better hallucination suppression.

\subsection{Qualitative Results}
\label{sec.QE}
\begin{figure*}[t]
\begin{center}
\includegraphics[width=1\linewidth]{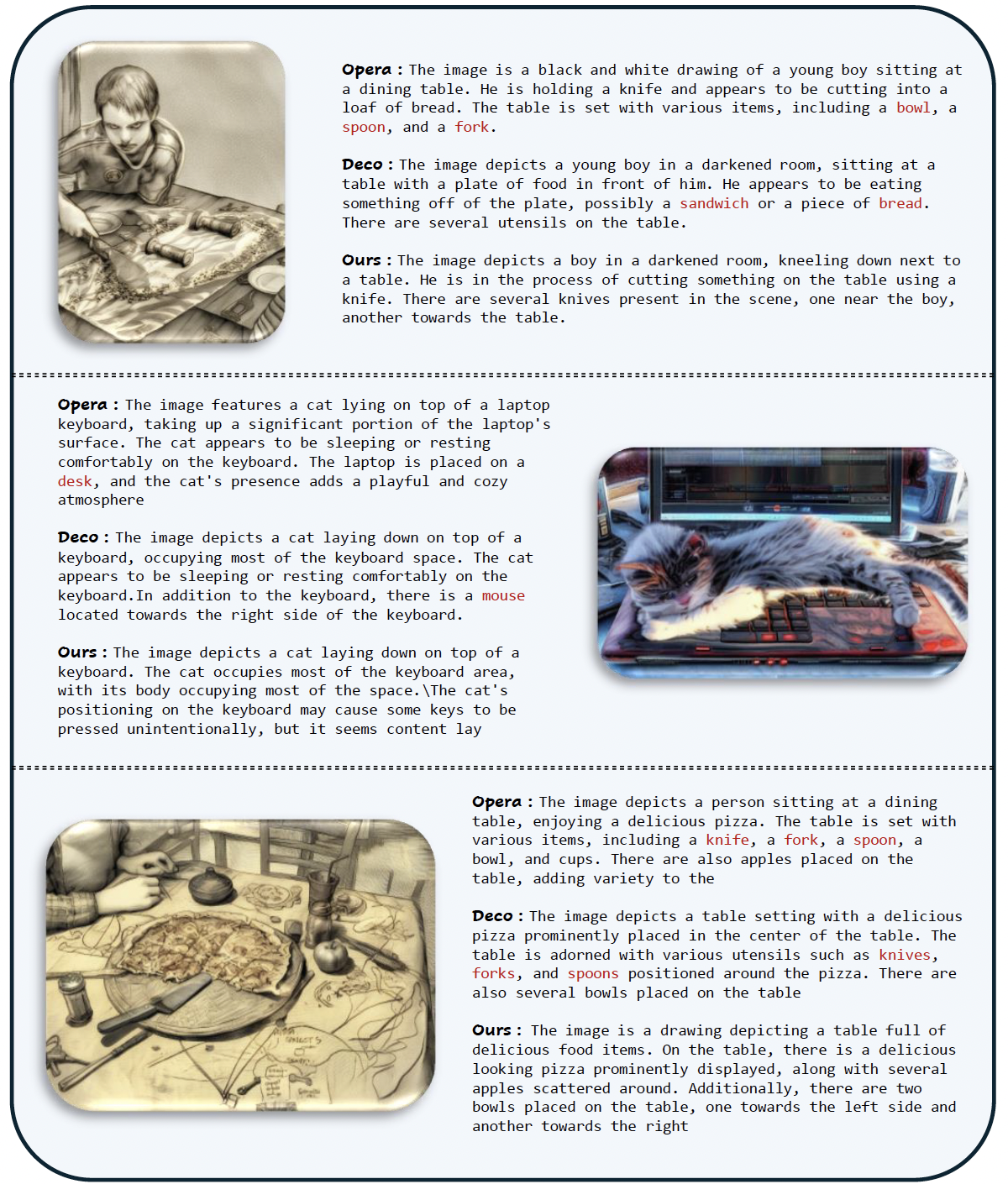}
\end{center}
   \caption{Qualitative results on LLaVA-1.5 using different decoding methods. Hallucinated objects are highlighted in red.}
\label{fig:example_llava}
\end{figure*}

\begin{figure*}[t]
\begin{center}
\includegraphics[width=1\linewidth]{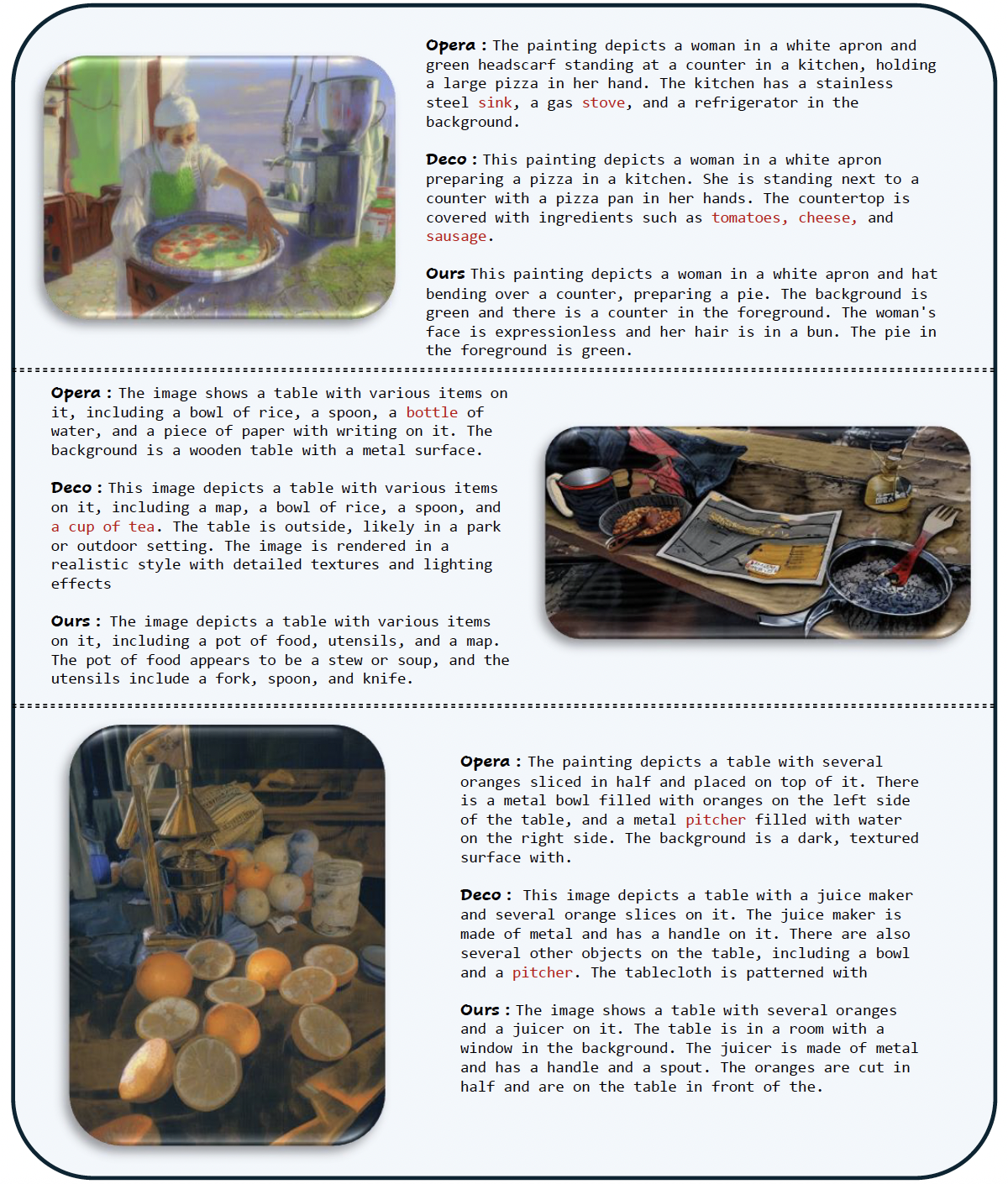}
\end{center}
   \caption{Qualitative results on MiniGPT-4 using different decoding methods. Hallucinated objects are highlighted in red.}
\label{fig:example_minigpt}
\end{figure*}

\begin{figure*}[t]
\begin{center}
\includegraphics[width=1\linewidth]{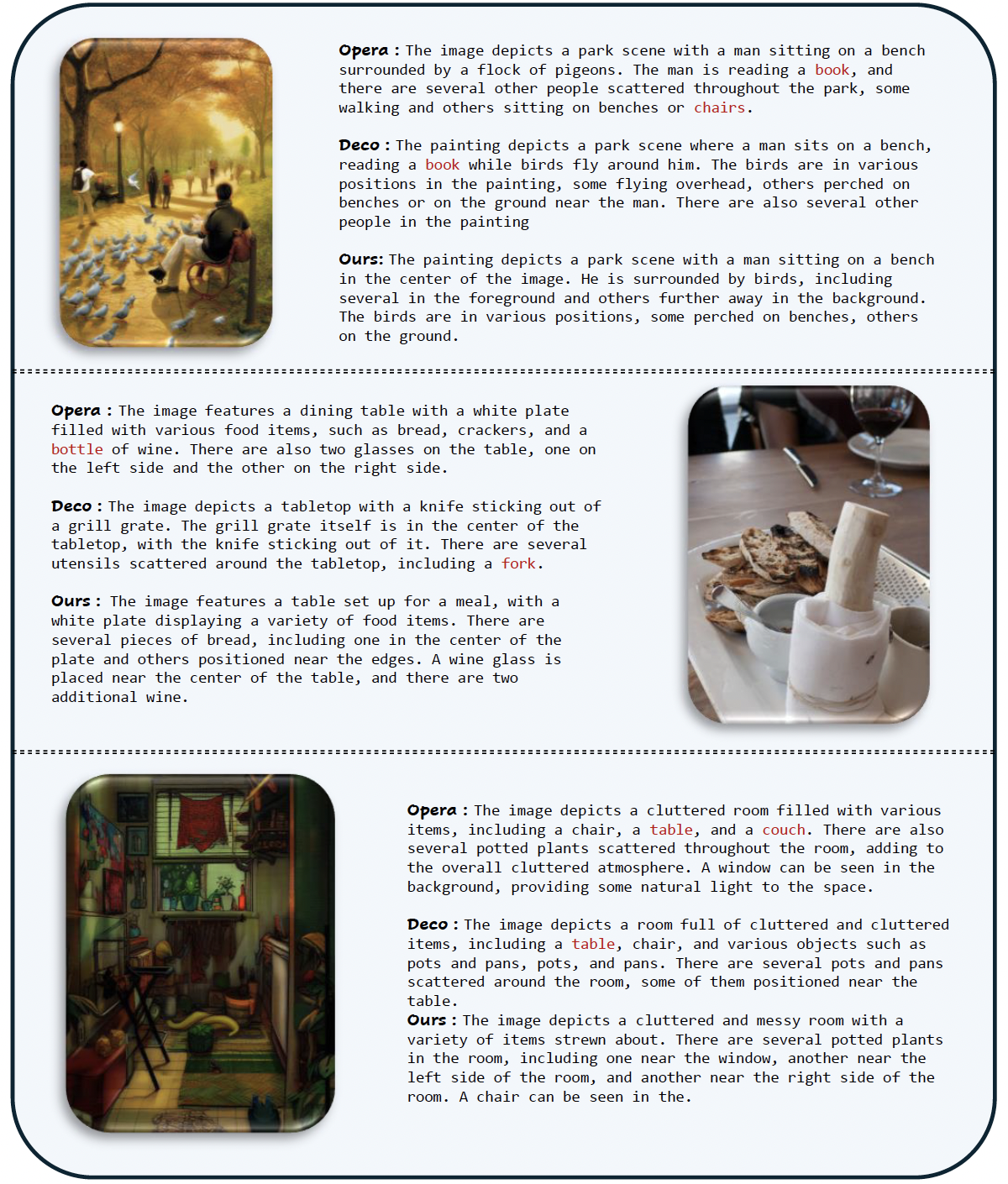}
\end{center}
   \caption{Qualitative results on InstructBLIP using different decoding methods. Hallucinated objects are highlighted in red.}
\label{fig:example_instructblip}
\end{figure*}

To qualitatively assess the responses produced by different models and decoding strategies, we present qualitative examples in Figs.~\ref{fig:example_llava},~\ref{fig:example_minigpt}, and~\ref{fig:example_instructblip}. Compared to SOTA methods, \textbf{SAVER} generates captions that maintain more descriptive detail and length while remarkably reducing object hallucinations.

\subsection{Hardware Setup}
\label{hardware}
We run our benchmark and decoding method comparison experiments using one NVIDIA A40 GPU with 48 GB memory.

\FloatBarrier
\begin{table*}[t]\small
\centering
\setlength{\tabcolsep}{4pt}

{%
\begin{tabular}{@{}c|cccc|cccccccccccc|cc@{}}
\toprule
\multirow{2}{*}[-0.3em]{Model}            & \multirow{2}{*}[-0.3em]{$\alpha$} & \multirow{2}{*}[-0.3em]{$N_i$} & \multirow{2}{*}[-0.3em]{$k$} & \multirow{2}{*}[-0.3em]{$p$} & \multicolumn{2}{c}{\textbf{Cartoon}} & \multicolumn{2}{c}{\textbf{Game}} & \multicolumn{2}{c}{\textbf{Graffiti}} & \multicolumn{2}{c}{\textbf{Painting}} & \multicolumn{2}{c}{\textbf{Sketch}} & \multicolumn{2}{c|}{\textbf{Original}} & \multicolumn{2}{c}{\textbf{Average}} \\ \cmidrule(l){6-19} 
                                  &                           &                        &                      &                      & Ci                & Cs               & Ci              & Cs              & Ci                & Cs                & Ci                & Cs                & Ci               & Cs               & Ci                & Cs                 & Ci               & Cs                \\ \midrule
\multirow{16}{*}[-0.8em]{LLaVA-1.5}    & 0.4                       & 50                     & 20                   & 0.9                  & 11.2              & 32.7             & \textbf{8.8}    & 29.7            & 11.4              & \textbf{30.3}     & 9.3               & \textbf{29.0}     & \textbf{10.6}    & \textbf{32.0}    & \textbf{6.6}      & 26.0               & \textbf{9.7}     & \textbf{30.0}     \\
                                  & 0.6                       & 50                     & 20                   & 0.9                  & \textbf{10.3}     & \textbf{32.0}    & 9.2             & 30.7            & \textbf{11.3}     & 30.7              & \textbf{9.1}      & 30.7              & 11.7             & 35.0             & 7.2               & 27.7               & 9.8              & 31.1              \\
\bottomrule
\end{tabular}%
}
\caption{Ablation studies on the hyper-parameters of SAVER with ``\textit{Standard}'' early exit layers, including the number of candidate tokens $k$, the scale factor $\alpha$, the number of candidate image representations $N_i$, and the threshold $p$. CHAIRs and CHAIRi are denoted as Cs and Ci. Lower Cs and Ci indicate fewer hallucinations. }
\label{tab:my-abl-1}
\end{table*}

\begin{table*}[t]\small
\centering
\setlength{\tabcolsep}{4pt}

{%
\begin{tabular}{@{}c|cccc|cccccccccccc|cc@{}}
\toprule
\multirow{2}{*}[-0.3em]{Model}            & \multirow{2}{*}[-0.3em]{$\alpha$} & \multirow{2}{*}[-0.3em]{$N_i$} & \multirow{2}{*}[-0.3em]{$k$} & \multirow{2}{*}[-0.3em]{$p$} & \multicolumn{2}{c}{\textbf{Cartoon}} & \multicolumn{2}{c}{\textbf{Game}} & \multicolumn{2}{c}{\textbf{Graffiti}} & \multicolumn{2}{c}{\textbf{Painting}} & \multicolumn{2}{c}{\textbf{Sketch}} & \multicolumn{2}{c|}{\textbf{Original}} & \multicolumn{2}{c}{\textbf{Average}}                     \\ \cmidrule(l){6-19} 
                                  &                           &                        &                      &                      & Ci                & Cs               & Ci              & Cs              & Ci                & Cs                & Ci                & Cs                & Ci               & Cs               & Ci                & Cs                 & \multicolumn{1}{c}{Ci} & \multicolumn{1}{c}{Cs} \\ \midrule
\multirow{16}{*}[-0.8em]{LLaVA-1.5}    & 0.4                       & 50                     & 20                   & 0.9                  & 10.8              & 36.3             & 9.6             & 33.7            & \textbf{10.5}     & 35.0              & 10.5              & 34.3              & \textbf{9.4}     & \textbf{32.0}    & 7.5               & 29.3               & 9.7                    & 33.4                   \\
                                  & 0.6                       & 50                     & 20                   & 0.9                  & 11.5              & 39.3             & 9.4             & 34.3            & 11.3              & 35.0              & 10.5              & 34.7              & 9.9              & \textbf{32.0}    & \textbf{7.1}      & \textbf{28.3}      & 10.0                   & 33.9                   \\
\bottomrule
\end{tabular}%
}
\caption{Ablation studies on the hyper-parameters of SAVER with ``\textit{Low}'' early exit layers, including the number of candidate tokens $k$, the scale factor $\alpha$, the number of candidate image representations $N_i$, and the threshold $p$. CHAIRs and CHAIRi are denoted as Cs and Ci. Lower Cs and Ci indicate fewer hallucinations.}
\label{tab:my-abl-2}
\end{table*}

\begin{table*}[t]\small
\centering
\setlength{\tabcolsep}{4pt}

{%
\begin{tabular}{@{}c|cccc|cccccccccccc|cc@{}}
\toprule
\multirow{2}{*}[-0.3em]{Model}            & \multirow{2}{*}[-0.3em]{$\alpha$} & \multirow{2}{*}[-0.3em]{$N_i$} & \multirow{2}{*}[-0.3em]{$k$} & \multirow{2}{*}[-0.3em]{$p$} & \multicolumn{2}{c}{\textbf{Cartoon}} & \multicolumn{2}{c}{\textbf{Game}} & \multicolumn{2}{c}{\textbf{Graffiti}} & \multicolumn{2}{c}{\textbf{Painting}} & \multicolumn{2}{c}{\textbf{Sketch}} & \multicolumn{2}{c|}{\textbf{Original}} & \multicolumn{2}{c}{\textbf{Average}} \\ \cmidrule(l){6-19} 
                                  &                           &                        &                      &                      & Ci                & Cs               & Ci              & Cs              & Ci                & Cs                & Ci                & Cs                & Ci               & Cs               & Ci                & Cs                 & Ci                & Cs               \\ \midrule
\multirow{16}{*}[-0.8em]{LLaVA-1.5}    & 0.4                       & 50                     & 20                   & 0.9                  & 11.3              & 37.0             & \textbf{8.2}    & \textbf{28.7}   & \textbf{12.1}     & \textbf{36.3}     & 11.2              & 35.7              & \textbf{9.5}     & 33.0             & \textbf{6.3}      & \textbf{24.0}      & \textbf{9.8}      & 32.5             \\
                                  & 0.6                       & 50                     & 20                   & 0.9                  & 11.3              & 37.0             & 8.3             & 29.7            & 12.3              & 37.3              & 11.3              & 33.3              & 9.8              & \textbf{32.3}    & 6.7               & 25.3               & 10.0              & 32.5             \\
\bottomrule
\end{tabular}%
}
\caption{Ablation studies on the hyper-parameters of SAVER with ``\textit{High}'' early exit layers, including the number of candidate tokens $k$, the scale factor $\alpha$, the number of candidate image representations $N_i$, and the threshold $p$. CHAIRs and CHAIRi are denoted as Cs and Ci. Lower Cs and Ci indicate fewer hallucinations.}
\label{tab:my-abl-3}
\end{table*}


\end{document}